\documentclass[10pt,twocolumn,letterpaper]{article}

\usepackage[pagenumbers]{wacv} % To force page numbers, e.g. for an arXiv version

\usepackage{graphicx}
\usepackage{amsmath}
\usepackage{amssymb}
\usepackage{booktabs}
\usepackage{multirow}
\usepackage[table,xcdraw]{xcolor}
\usepackage[normalem]{ulem}
\useunder{\uline}{\ul}{}
\usepackage{stmaryrd}
\usepackage[accsupp]{axessibility}  % Improves PDF readability for those with disabilities.

\usepackage[pagebackref,breaklinks,colorlinks]{hyperref}
\usepackage{appendix}

\usepackage[capitalize]{cleveref}
\crefname{section}{Sec.}{Secs.}
\Crefname{section}{Section}{Sections}
\Crefname{table}{Table}{Tables}
\crefname{table}{Tab.}{Tabs.}

\definecolor{higher_better}{HTML}{94BBE8}
\definecolor{lower_better}{HTML}{E8A194}

\def\wacvPaperID{2052} % *** Enter the WACV Paper ID here
\def\confName{WACV}
\def\confYear{2024}

\begin{document}

\title{MonoProb: Self-Supervised Monocular Depth Estimation with Interpretable Uncertainty}

\author{
Rémi MARSAL\footnote[1]{} \footnote[2]{} \and Florian CHABOT\footnote[1]{} \and Angelique LOESCH\footnote[1]{} \and William GROLLEAU\footnote[1]{} \and Hichem SAHBI\footnote[2]{}\\
\footnote[1]{}  Université Paris-Saclay, CEA, LIST, F-91120, Palaiseau, France \\
{\tt\small firstname.lastname@cea.fr} \\
\footnote[2]{}  Sorbonne University, CNRS, LIP6 F-75005, Paris, France \\
{\tt\small firstname.lastname@lip6.fr}
}

\maketitle

%%%%%%%%% ABSTRACT
\begin{abstract}
Self-supervised monocular depth estimation methods aim to be used in critical applications such as autonomous vehicles for environment analysis. To circumvent the potential imperfections of these approaches, a quantification of the prediction confidence is crucial to guide decision-making systems that rely on depth estimation. In this paper, we propose MonoProb, a new unsupervised monocular depth estimation method that returns an interpretable uncertainty, which means that the uncertainty reflects the expected error of the network in its depth predictions. We rethink the stereo or the structure-from-motion paradigms used to train unsupervised monocular depth models as a probabilistic problem. Within a single forward pass inference, this model provides a depth prediction and a measure of its confidence, without increasing the inference time. We then improve the performance on depth and uncertainty with a novel self-distillation loss for which a student is supervised by a pseudo ground truth that is a probability distribution on depth output by a teacher. To quantify the performance of our models we design new metrics that, unlike traditional ones, measure the absolute performance of uncertainty predictions. Our experiments highlight enhancements achieved by our method on standard depth and uncertainty metrics as well as on our tailored metrics. \href{https://github.com/CEA-LIST/MonoProb}{\texttt{https://github.com/CEA-LIST/MonoProb}}
\end{abstract}

%------------------------------------------------------------------------
\section{Introduction}

Advances in deep learning in the field of computer vision have led to breakthroughs in depth estimation \cite{eigen2014depth, liu2015deep}. This task is crucial for applications like autonomous driving, as it provides an analysis of the environment that can inform about the presence of obstacles, for instance. Therefore, it must be sufficiently reliable to be used for decision-making. In particular, the expected common criteria are high performance, fast inference, and the ability to quantify the confidence in the model predictions. However, these requirements often conflict with the inherent limitations of traditional deep learning methods. Indeed, these approaches require extensive labeled datasets, incurring substantial additional costs. Furthermore, they are black-box systems, yielding predictions without reliability clue. Considering supervised learning for monocular depth estimation, labeled data can be provided by recording videos of multiple scenes while performing a synchronized lidar acquisition at the same time \cite{kitti, nuscenes}. An alternative consists in training a  model on synthetic data \cite{packnet} which nevertheless introduces a domain gap between the training set and real use-case data.

The challenge of acquiring labeled data can be mitigated through unsupervised training strategies, where the objective is to minimize the reconstruction error between source and target images captured from a different perspective \cite{garg2016unsupervised, godard2017unsupervised} or at a different instant \cite{zhou2017unsupervised} within the same scene. The lack of confidence can be addressed by ensembling methods \cite{kendalluncert, poggi2020uncertainty} that combine the results of multiple inferences from one or more models to obtain a variance for each prediction. However, the aforementioned approaches incur computational overhead during both training and inference, particularly in the context of bootstrap ensembles. Alternatively, predictive methods \cite{kendalluncert, klodt2018supervising, poggi2020uncertainty} return complementary outputs in addition to the depth map to quantify the uncertainty. Since they require only one inference per frame, they are more attractive for real-time applications.

In the case of supervised learning, a simple and direct uncertainty estimation approach is to model depth with a probability distribution \cite{kendalluncert}. The estimator returns the parameters of this distribution instead of scalars and the traditional distance to the ground truth as loss function is replaced by the likelihood maximization of the predicted depth distribution. Notably, this method offers the advantage of providing interpretable uncertainty values within a single inference that is the depth variance. In this paper, we consider that an uncertainty is interpretable if it gives an estimate of the expected error between the predicted depth and the ground truth. However, this technique depends on the availability of ground truth data. In the context of unsupervised learning, several works \cite{klodt2018supervising, poggi2020uncertainty} adapt this approach by modeling the image reconstruction by a probability distribution and maximizing the likelihood of the reconstruction. By leveraging the weight of pixels where the supervisory signal from the reconstruction loss is unreliable, they achieve performance enhancements. Still, this uncertainty is difficult to use in practical applications due to its complex link with depth that limits its interpretability.

In this paper, (1) we first present MonoProb, an unsupervised depth estimation method that provides an interpretable confidence measure of its predictions. This approach extends the likelihood maximization strategy to unsupervised learning. Unlike \cite{klodt2018supervising, poggi2020uncertainty} that only model the reconstruction by a probability distribution, we express this probabilistic reconstruction with respect to a probability distribution over depth. As a result, this technique provides an interpretable and reliable uncertainty relative to depth in a single inference, enabling informed subsequent decision-making in real-time. This uncertainty is the standard deviation (STD) of a predicted depth distribution. Furthermore, we demonstrate the ability of MonoProb to improve performance on depth estimation. (2) Second, \cite{poggi2020uncertainty} estimates uncertainty using a self-distillation method where pseudo ground truth scalar depth maps from a teacher model supervise a student model. We enhance the quality of our uncertainty predictions by adapting this approach to handle probability distributions as pseudo ground truth provided by the teacher. (3) Finally, we propose two new metrics tailored for evaluating the quality of interpretable uncertainty. To compare different types of uncertainties, specifically non-interpretable ones, \cite{ilg2018uncertainty, poggi2020uncertainty} introduced metrics that are invariant to an increasing bijection on uncertainty. They measure a relative uncertainty within an image by indicating the effectiveness of uncertainty in sorting image pixels by order of performance on a given depth metric. Conversely, our metrics are designed by considering the ability of the network to predict its absolute performance. 
% Finally, our contributions can be summarized as follows:
% \begin{itemize}
%     \item MonoProb, an unsupervised depth estimation method that provides an interpretable uncertainty,
%     \item a loss for the self-distillation of depth distributions as pseudo ground truth,
%     \item new metrics to quantify absolute uncertainty.
% \end{itemize}

% Experiments on KITTI \cite{kitti} show that our method improve performances 

%------------------------------------------------------------------------
\section{Related works}

\begin{figure*}[h]
    \centering
    \includegraphics[]{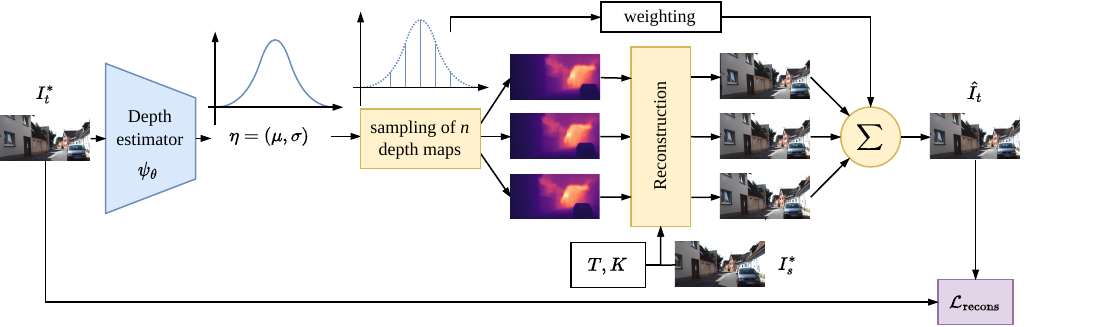}
    \caption{Our depth estimator takes the target image $I_t^*$ as input and returns a map of the parameters $\eta=(\mu, \sigma)$ of a multi-variate depth distribution $D$. A sampling yields $n$ depth maps used to reconstruct $n$ times $I_t^*$. These reconstructions also involve a source image $I_s^*$, the intrinsic camera calibration $K$, and the camera motion $T$ between $I_s^*$ and $I_t^*$. They are then weighted according to the parameters $\eta$ of the distribution $D$ and averaged. This gives the final reconstruction $\hat{I}_t$, which is compared to the original $I_t^*$ in the $\mathcal{L}_{\text{recons}}$ loss.}
    \label{fig:diagram}
\end{figure*}

\subsection{Unsupervised monocular depth estimation}

When no ground truth is available, image reconstruction stands as a widely popular pretext task. It enables the model to learn depth by linking it to the apparent motion of pixels characterizing the same object between two images of a scene in a unique way. This technique also requires the camera's intrinsic calibration and the relative position of the camera between two views of a scene. Pioneering works use either stereo images \cite{garg2016unsupervised,godard2017unsupervised} or image sequences \cite{zhou2017unsupervised} for this purpose, then \cite{godard2019digging, poggi2018learning, zhan2018unsupervised} combine them both at training. Subsequent works improve performance by guiding the model training with traditional non-learning based methods \cite{klodt2018supervising, watson2019self}, using learned features \cite{shu2020feature}, enhancing network architecture design, incorporating semantic guidance \cite{chen2019towards, klingner2020self, guizilini20203d} or regularizing training with self-distillation \cite{pilzer2019refine, peng2021excavating, bello2021self}.

\subsection{Uncertainty for supervised learning}

The study of uncertainty estimation for depth estimation task in deep learning has gained significant attention in recent years. These methods can be categorized into two families: ensembling and predictive methods. Ensembling methods assume that neural network weights follow a probability distribution. Uncertainty is then given by the variance of multiple predictions obtained by sampling several model weights. Bayesian neural networks \cite{chen2014stochastic, mackay1992practical, welling2011bayesian} explicitly assign a probability distribution to each individual weight. Monte Carlo dropout \cite{gal2016dropout} consists in randomly switching off network weights. Bootstrapped ensembles \cite{lakshminarayanan2017simple} involve training several models on different subsets of the training dataset, while in Snapshot Ensembles \cite{huang2017snapshot} store multiple models from a single training. On the other hand, predictive methods \cite{kendalluncert} aim at returning direct quantification of the uncertainty in their outputs. These methods are less computationally intensive, requiring only a single inference step, although they require an adjustment of the loss function. The outputs belong to a family of distributions whose parameters are estimated by a neural network. The loss function to be minimized is the negative log-likelihood (NLL) of the actual ground truth, making it incompatible for direct use in unsupervised learning scenarios. In our work, we adapt this approach for the unsupervised learning of a probabilistic depth distribution. 

% This general method is applied to learn various unsupervised tasks such as optical flow, depth or ego-motion. It results that they share common challenges such as occlusions, brightness changes or uniform surfaces. Likewise, the issue of uncertainty could be tackled in a similar way for all these tasks. However, previous works tend to treat each of these tasks separately.

% \paragraph{Uncertainty for optical flow estimation} 

% The issue of uncertainty in optical flow has been a subject of extensive investigation in the literature. However, to the best of our knowledge, \cite{ilg2018uncertainty} is the only work that estimates uncertainty for deep learning optical flow network. It compares several empirical ensembling methods, predictive methods, the combination of both and a multi-hypothesis network. 

\subsection{Uncertainty for unsupervised depth estimation}

Early unsupervised monocular depth estimation methods include uncertainty reconstruction. The explainability mask of Zhou et al.'s method \cite{zhou2017unsupervised} encodes the network's confidence in its ability to reconstruct each pixel. Since it is used to weight each pixel of the reconstruction loss differently, it needs a regularization term to prevent a trivial solution. \cite{klodt2018supervising, yang2020d3vo}, inspired by \cite{kendalluncert}, remove the regulation term and incorporate the uncertainty map in a likelihood maximization-like problem of reconstruction where the reconstruction error is modeled by a Laplacian distribution. Thus, the predicted variance quantifies how likely the reconstruction loss is to be minimized. However, this is not exactly equivalent to uncertainty on depth because on photorealistic datasets \cite{kitti, nuscenes, make3d}, even a perfectly estimated depth map can produce reconstruction failures due to phenomena like brightness changes, non-Lambertian surfaces or highly detailed areas. Nevertheless, it can serve as a useful tool for approximating depth uncertainty as shown by Poggi et al. \cite{poggi2020uncertainty}. They provide a broad synthesis of techniques that can be used to estimate uncertainty, ranging from empirical approaches such as dropout or ensembling methods to predictive approaches such as likelihood maximization of the reconstruction loss or of a pseudo ground truth, or the prediction of the reconstruction loss. They also combine empirical and predictive approaches. Several works \cite{johnston2020self, gonzalezbello2020forget}. reformulate depth estimation into a classification problem of discrete disparities that can be used to estimate a variance. Dikov et al. \cite{dikov2022variational} propose the first predictive approach of an interpretable uncertainty that is depth STD. The reconstruction loss is reformulated as a reconstruction likelihood expressed with respect to the depth distribution thanks to Bayes' theorem. Instead, we chose the law of total probability to make the depth distribution appear in the reconstruction likelihood. This enables to avoid learning pseudo-inputs as in \cite{dikov2022variational} to get a prior distribution on depth. We demonstrate that this improves both depth and uncertainty performance. 
% Empirical results show that our method improves both depth and uncertainty performance relative to \cite{dikov2022variational} (see \cref{tab:res_VDN}).
%------------------------------------------------------------------------

\section{Unsupervised depth estimation}
\label{sec:unsup_depth}

Consider a collection of image pairs denoted as $\mathcal{I}=\{(I^*_s, I^*_t)_i\}_i$ where each image $I^*_j$, $j \in \{s, t\}$ in a pair abides by $I^*_j\in \mathbb{R}^{H\times W \times C}$ and $H$, $W$, $C$ respectively stand for the height, the width and the number of channels. We assume each image of a pair represents a different point of view of a same scene which verifies the brightness consistency assumption and has no uniformly textured surface. The change of point of view can result from a small movement of the camera relative to the scene (e.g. stereo images) or from a different timestamp (as in structure-from-motion methods \cite{vijayanarasimhan2017sfm}). The brightness consistency assumption means that objects in a scene keep a constant brightness between two different close views. Hence, pixel motion is sufficient to explain the transformation to be applied to an image named the source image $I_s^*$ of an image pair in $\mathcal{I}$ to obtain the other one called the target image $I_t^*$. In a static scene, the new location $p_s$ in $I_s^*$ of a pixel $p_t$ from $I_t^*$ can be expressed as a function of the depth $D_t$ of the target image, the camera motion $T_{t\rightarrow s}$ from the target to the source image and the camera calibration $K$ with the formula:
\begin{equation}
    p_s = K T_{t\rightarrow s} D_t(p_t) K^{-1} p_t.
\end{equation}

For each pixel $p_t$, sampling in the source image at location $p_s$ results in a new image that is a reconstruction of the target image $I_t^*$. Thus, minimizing the reconstruction error provides a relevant pretext task for the unsupervised learning of depth, given camera motion and calibration.

\section{Method}

\begin{table*}[]
\scriptsize
\centering
\begin{tabular}{|c|c|c|c|c|c|cc|cc|cc|c|c|}
\hline
 &  &  & \cellcolor[HTML]{E8A194} & \cellcolor[HTML]{E8A194} & \cellcolor[HTML]{94BBE8} & \multicolumn{2}{c|}{Abs Rel} & \multicolumn{2}{c|}{RMSE} & \multicolumn{2}{c|}{$\delta < 1.25$} & \cellcolor[HTML]{E8A194} & \cellcolor[HTML]{E8A194} \\
\multirow{-2}{*}{Sup} & \multirow{-2}{*}{Resolution} & \multirow{-2}{*}{\#Trn} & \multirow{-2}{*}{\cellcolor[HTML]{E8A194}Abs Rel $\downarrow$} & \multirow{-2}{*}{\cellcolor[HTML]{E8A194}RMSE $\downarrow$} & \multirow{-2}{*}{\cellcolor[HTML]{94BBE8}$\delta < 1.25$ $\uparrow$} & \multicolumn{1}{c|}{\cellcolor[HTML]{E8A194}AUSE $\downarrow$} & \cellcolor[HTML]{94BBE8}AURG$\uparrow$ & \multicolumn{1}{c|}{\cellcolor[HTML]{E8A194}AUSE $\downarrow$} & \cellcolor[HTML]{94BBE8}AURG$\uparrow$ & \multicolumn{1}{c|}{\cellcolor[HTML]{E8A194}AUSE $\downarrow$} & \cellcolor[HTML]{94BBE8}AURG$\uparrow$ & \multirow{-2}{*}{\cellcolor[HTML]{E8A194}ARU$\downarrow$} & \multirow{-2}{*}{\cellcolor[HTML]{E8A194}RMSU$\downarrow$} \\ \hline
M & \cite{godard2019digging} & 1 & {\ul 0.090} & 3.942 & \textbf{0.914} & \multicolumn{1}{c|}{-} & - & \multicolumn{1}{c|}{-} & - & \multicolumn{1}{c|}{-} & - & - & - \\
M & \cite{poggi2020uncertainty}-Repr & 1 & 0.092 & {\ul 3.936} & {\ul 0.912} & \multicolumn{1}{c|}{0.051} & 0.008 & \multicolumn{1}{c|}{2.972} & 0.381 & \multicolumn{1}{c|}{0.069} & 0.013 & - & - \\
M & \cite{poggi2020uncertainty}-Log & 1 & 0.091 & 4.052 & 0.910 & \multicolumn{1}{c|}{{\ul 0.039}} & {\ul 0.020} & \multicolumn{1}{c|}{{\ul 2.562}} & {\ul 0.916} & \multicolumn{1}{c|}{{\ul 0.044}} & {\ul 0.038} & - & - \\
M & \textbf{Ours} & 1 & \textbf{0.089} & \textbf{3.852} & \textbf{0.914} & \multicolumn{1}{c|}{\textbf{0.031}} & \textbf{0.026} & \multicolumn{1}{c|}{\textbf{0.719}} & \textbf{2.560} & \multicolumn{1}{c|}{\textbf{0.030}} & \textbf{0.050} & \textbf{0.064} & \textbf{2.912} \\ \hline
M & \cite{poggi2020uncertainty}-Self & 2 & \textbf{0.087} & 3.826 & \textbf{0.920} & \multicolumn{1}{c|}{0.030} & 0.026 & \multicolumn{1}{c|}{2.009} & 1.266 & \multicolumn{1}{c|}{0.030} & 0.045 & 0.074 & 3.730 \\
M & \textbf{Ours-self} & 2 & \textbf{0.087} & \textbf{3.762} & 0.919 & \multicolumn{1}{c|}{\textbf{0.022}} & \textbf{0.034} & \multicolumn{1}{c|}{\textbf{0.326}} & \textbf{2.880} & \multicolumn{1}{c|}{\textbf{0.014}} & \textbf{0.061} & \textbf{0.066} & \textbf{2.969} \\ \hline
S & \cite{godard2019digging} & 1 & {\ul 0.085} & 3.942 & 0.912 & \multicolumn{1}{c|}{-} & - & \multicolumn{1}{c|}{-} & - & \multicolumn{1}{c|}{-} & - & - & - \\
S & \cite{poggi2020uncertainty}-Repr & 1 & {\ul 0.085} & 3.873 & 0.913 & \multicolumn{1}{c|}{0.040} & 0.017 & \multicolumn{1}{c|}{2.275} & 1.074 & \multicolumn{1}{c|}{0.050} & 0.030 & - & - \\
S & \cite{poggi2020uncertainty}-Log & 1 & {\ul 0.085} & {\ul 3.860} & {\ul 0.915} & \multicolumn{1}{c|}{\textbf{0.022}} & \textbf{0.036} & \multicolumn{1}{c|}{{\ul 0.938}} & {\ul 2.402} & \multicolumn{1}{c|}{\textbf{0.018}} & \textbf{0.061} & - & - \\
S & \textbf{Ours} & 1 & \textbf{0.084} & \textbf{3.834} & \textbf{0.916} & \multicolumn{1}{c|}{{\ul 0.023}} & {\ul 0.033} & \multicolumn{1}{c|}{\textbf{0.661}} & \textbf{2.655} & \multicolumn{1}{c|}{{\ul 0.023}} & {\ul 0.055} & \textbf{0.075} & \textbf{3.540} \\ \hline
S & \cite{poggi2020uncertainty}-Self & 2 & \textbf{0.084} & 3.835 & \textbf{0.915} & \multicolumn{1}{c|}{0.022} & 0.035 & \multicolumn{1}{c|}{1.679} & 1.642 & \multicolumn{1}{c|}{0.022} & 0.056 & 0.083 & 3.686 \\
S & \textbf{Ours-self} & 2 & \textbf{0.084} & \textbf{3.792} & 0.914 & \multicolumn{1}{c|}{\textbf{0.018}} & \textbf{0.038} & \multicolumn{1}{c|}{\textbf{0.349}} & \textbf{2.924} & \multicolumn{1}{c|}{\textbf{0.019}} & \textbf{0.060} & \textbf{0.072} & \textbf{3.068} \\ \hline
MS & \cite{godard2019digging} & 1 & {\ul 0.084} & \textbf{3.739} & \textbf{0.918} & \multicolumn{1}{c|}{-} & - & \multicolumn{1}{c|}{-} & - & \multicolumn{1}{c|}{-} & - & - & - \\
MS & \cite{poggi2020uncertainty}-Repr & 1 & {\ul 0.084} & 3.828 & 0.913 & \multicolumn{1}{c|}{0.046} & {\ul 0.010} & \multicolumn{1}{c|}{2.662} & 0.635 & \multicolumn{1}{c|}{0.062} & 0.018 & - & - \\
MS & \cite{poggi2020uncertainty}-Log & 1 & \textbf{0.083} & {\ul 3.790} & {\ul 0.916} & \multicolumn{1}{c|}{{\ul 0.028}} & \textbf{0.029} & \multicolumn{1}{c|}{{\ul 1.714}} & {\ul 1.562} & \multicolumn{1}{c|}{\textbf{0.028}} & \textbf{0.050} & - & - \\
MS & \textbf{Ours} & 1 & {\ul 0.084} & 3.806 & 0.915 & \multicolumn{1}{c|}{\textbf{0.027}} & \textbf{0.029} & \multicolumn{1}{c|}{\textbf{0.840}} & \textbf{2.436} & \multicolumn{1}{c|}{{\ul 0.029}} & {\ul 0.049} & \textbf{0.077} & \textbf{3.573} \\ \hline
MS & \cite{poggi2020uncertainty}-Self & 2 & 0.083 & 3.682 & \textbf{0.919} & \multicolumn{1}{c|}{0.022} & 0.033 & \multicolumn{1}{c|}{1.654} & 1.515 & \multicolumn{1}{c|}{0.023} & 0.052 & 0.083 & 3.686 \\
MS & \textbf{Ours-self} & 2 & \textbf{0.082} & \textbf{3.667} & \textbf{0.919} & \multicolumn{1}{c|}{\textbf{0.016}} & \textbf{0.039} & \multicolumn{1}{c|}{\textbf{0.293}} & \textbf{2.859} & \multicolumn{1}{c|}{\textbf{0.014}} & \textbf{0.061} & \textbf{0.078} & \textbf{3.528} \\ \hline
\end{tabular}
\caption{Results of monocular only (M), stereo only (S) and monocular and stereo (MS) trainings of our MonoProb with and without self-distillation compared to other methods.}
\label{tab:res}
\end{table*}

\subsection{Reconstruction loss with probabilistic depth}

We consider a depth uncertainty to be interpretable if its value directly informs about the expected depth error. To provide depth with an interpretable uncertainty, we choose to model depth by a distribution $D$ whose mean $\mu_D$ is used as depth prediction and STD $\sigma_D$ as uncertainty prediction. The unsupervised learning framework described in \cref{sec:unsup_depth} is re-designed into a probabilistic paradigm. For this purpose, the image reconstruction is modeled by a distribution that is expressed as a function of the depth distribution $D$ by applying the law of total probability. Since this formulation requires the costly computation of an integral over depth, we propose several approximations to make it tractable. This leads to a reformulation of the reconstruction loss. Our method is illustrated in \cref{fig:diagram}.

\paragraph{Probabilistic model.}

The reconstruction of the target image is modeled by a distribution $I_t$. Using the law of total probability, the likelihood of the reconstruction conditionally on $I_s^*$, $I\mapsto \text{p}_{I_t}(I|I_s^*)$ can be expressed with respect to depth distribution $D$ defined on $\mathcal{D}$ so that:
\begin{equation}
    \text{p}_{I_t}(I\:|\:I_s^*) = \int_{\mathcal{D}} \text{p}_{I_t}(I\:|\:d, I_s^*)\: \text{p}_D(d)\: \text{d}d.
\end{equation}

Given $\text{p}_{I_t}(I|d, I_s^*)$, maximizing the likelihood $\text{p}_{I_t}(I|I_s^*)$ with respect to $\text{p}_D$ for $I=I_t^*$ provides an estimator of $D$.

\paragraph{Neural network estimator.}

The depth distribution $D$ is assumed to belong to a family of distributions $\mathcal{L}_{\eta}$ with unknown parameters $\eta$. We introduce $\psi_{\theta}$ a function with learnable parameters $\theta$ that returns the $\eta$ parameters of  $D$ given an image pair $(I_t^*, I_s^*)\in \mathcal{I}$. The likelihood of the reconstruction becomes:
\begin{equation}
    \label{eq:2}
        \text{p}_{I_t}(I\:|\:I_s^*, \theta) = \int_{\mathcal{D}} \text{p}_{I_t}(I\:|\:d, I_s^*) \text{p}_D(d\:|\:\eta) \text{d}d,
\end{equation}

\noindent with $\eta = \psi_{\theta}(I_t^*)$. Thus, $\psi_{\theta}$ is an estimator of $D$ by minimizing the negative log-likelihood of the reconstruction with respect to the parameters $\theta$ when $I=I_t^*$. Let $\text{recons}(., .)$ be a reconstruction function that takes as input a punctual estimate of depth and an image, and returns an image so that $\text{recons}(D^*, I_s^*)=I_t^*$ and $\text{err}(.,.)$ a distance between two images, we define $\text{p}_{I_t}(I\:|\:d, I_s^*) = \frac{1}{\lambda} \exp(-\text{err}(\text{recons}(d, I_s^*), I))$. The parameter $\lambda$ aims at enforcing the upper bound of the cumulative distribution function relative to $\text{p}_{I_t}(I|d, I_s^*)$ to be equal to $1$, thus $\lambda = \int_{\mathcal{D}} \exp(-\text{err}(\text{recons}(d, I_s^*), I))\text{d}d$. The choice for $\text{p}_{I_t}(I\:|\:d, I_s^*)$ enables to recover \cite{godard2019digging}'s reconstruction error when computing the negative log-likelihood of $\text{p}_{I_t}(I\:|\:I_s^*)$ with a punctual distribution as $D$ distribution.

\begin{figure*}[h]
    \centering
    \includegraphics[width=1.01\textwidth]{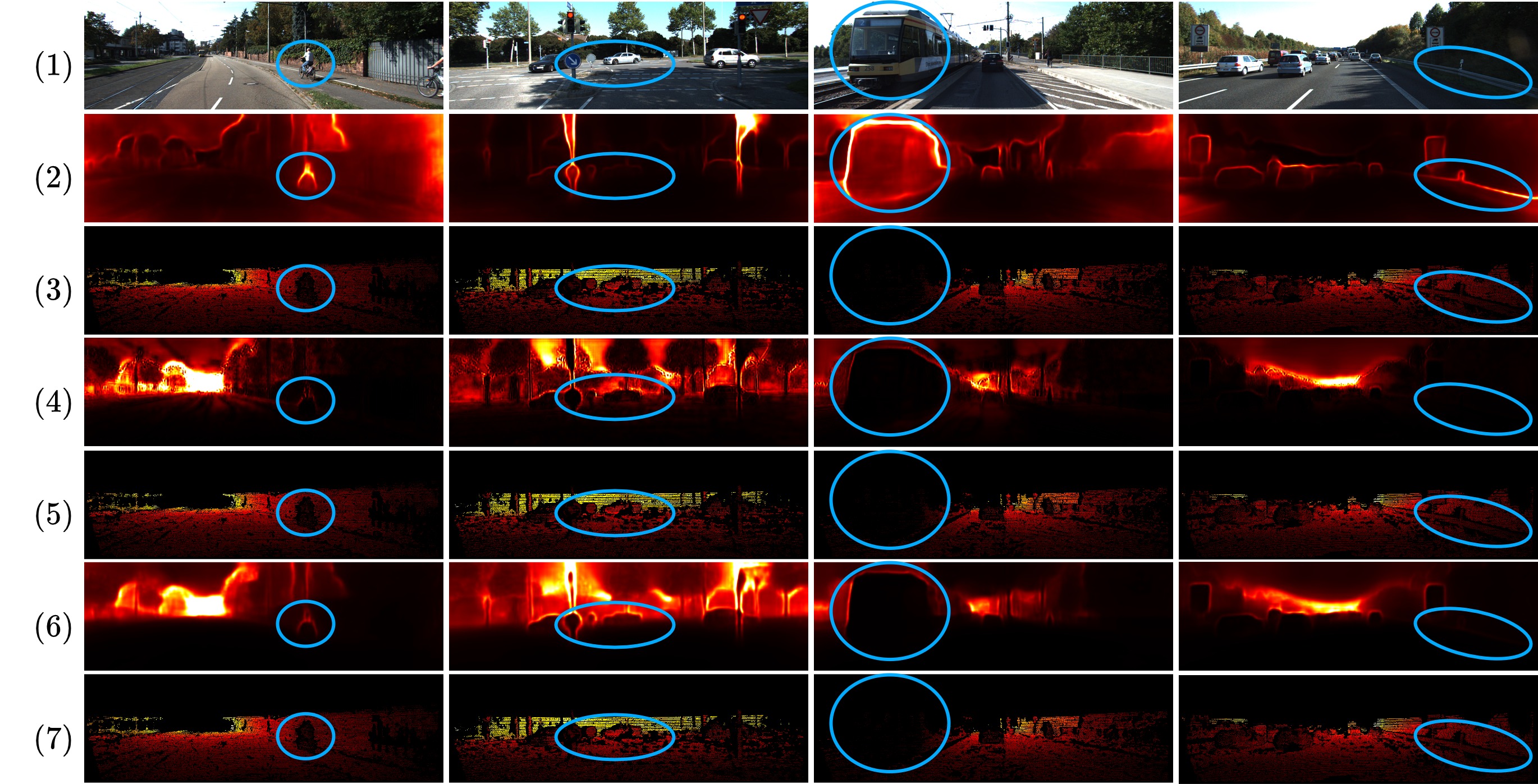}
    \caption{Qualitative results of trainings on KITTI monocular videos. (1) Input image, (2) Uncertainty map from \cite{poggi2020uncertainty}-Self, (3) Depth error map from \cite{poggi2020uncertainty}-Self, (4) Uncertainty map from our MonoProb without self-distillation, (5) Depth error map from our MonoProb without self-distillation, (6) Uncertainty map from our  MonoProb with self-distillation, (7) Depth error map from our self-distilled MonoProb.}
    \label{fig:visu}
\end{figure*}

\begin{table*}[]
\centering
\scriptsize
\begin{tabular}{|c|c|c|c|cc|cc|cc|}
\hline
 & \cellcolor[HTML]{E8A194} & \cellcolor[HTML]{E8A194} & \cellcolor[HTML]{94BBE8} & \multicolumn{2}{c|}{Abs Rel} & \multicolumn{2}{c|}{RMSE} & \multicolumn{2}{c|}{$\delta < 1.25$} \\ \cline{5-10} 
\multirow{-2}{*}{Methods} & \multirow{-2}{*}{\cellcolor[HTML]{E8A194}Abs Rel $\downarrow$} & \multirow{-2}{*}{\cellcolor[HTML]{E8A194}RMSE $\downarrow$} & \multirow{-2}{*}{\cellcolor[HTML]{94BBE8}$\delta < 1.25$ $\uparrow$} & \multicolumn{1}{c|}{\cellcolor[HTML]{E8A194}AUSE $\downarrow$} & \cellcolor[HTML]{94BBE8}AURG$\uparrow$ & \multicolumn{1}{c|}{\cellcolor[HTML]{E8A194}AUSE $\downarrow$} & \cellcolor[HTML]{94BBE8}AURG$\uparrow$ & \multicolumn{1}{c|}{\cellcolor[HTML]{E8A194}AUSE $\downarrow$} & \cellcolor[HTML]{94BBE8}AURG$\uparrow$ \\ \hline
VDN \cite{dikov2022variational} & 0.117 & 4.815 & 0.873 & \multicolumn{1}{c|}{0.058} & 0.018 & \multicolumn{1}{c|}{1.942} & 2.140 & \multicolumn{1}{c|}{0.085} & 0.030 \\
\textbf{Ours} & \textbf{0.114} & \textbf{4.772} & \textbf{0.875} & \multicolumn{1}{c|}{\textbf{0.044}} & \textbf{0.030} & \multicolumn{1}{c|}{\textbf{1.625}} & \textbf{2.437} & \multicolumn{1}{c|}{\textbf{0.060}} & \textbf{0.054} \\ \hline
\end{tabular}
\caption{Comparison with VDN \cite{dikov2022variational} on KITTI with the raw ground truth.}
\label{tab:res_VDN}
\end{table*}

\paragraph{Approximations of the reconstruction likelihood.}

In this paper, $\psi_{\theta}$ is a neural network that returns a map $\eta$ of the parameters of the distribution $D \in \mathcal{D} = \mathbb{R}^{H\times W\times 1}$ of depth. To make the prediction of $\eta$ tractable with usual state-of-the-art neural networks, $\psi_{\theta}$ outputs are restricted to the parameters of the $H\times W\times 1$ marginal distributions of $D$. This means that $\text{p}_D(d|\eta)$ in \cref{eq:2} is partially unknown. A solution to this problem is to minimize an approximation to the upper bound of the negative log-likelihood of the reconstruction rather than the negative log-likelihood itself. Thus, using the convex property of the $x\mapsto-\log(x)$ function, the Jensen inequality can be applied as follows:
\begin{equation}
    \begin{split}
        -\log \text{p}_{I_t}(I\:|\:I_s^*, \theta) & = -\log\int_{\mathcal{D}} \text{p}_{I_t}(I\:|\:d, I_s^*) \text{p}_D(d\:|\:\eta) \text{d}d \\
        & = -\log \mathbb{E}_D[\text{p}_{I_t}(I\:|\:D, I_s^*)\:|\:\eta] \\
        & \leq \mathbb{E}_D[-\log \text{p}_{I_t}(I\:|\:D, I_s^*)\:|\:\eta].
    \end{split}
\end{equation}

Then this upper bound is approximated:
\begin{equation}
    \begin{split}
        \mathbb{E}_D[-\log \text{p}_{I_t}(I|D, I_s^*)|\eta] & = \mathbb{E}_D[\text{err}(\text{recons}(D, I_s^*), I)|\eta] + c \\
        & \approx \text{err}(\mathbb{E}_D[\text{recons}(D, I_s^*)|\eta], I) + c.
    \end{split}
    \label{eq:approx}
\end{equation}

\noindent where $c=-\log(\lambda)$ is a constant. The case of equality occurs when the depth estimator $\psi_{\theta}$ returns a punctual distribution, i.e. with a null scale parameter. This means that the model is absolutely certain about its prediction. This is a behavior that $\psi_{\theta}$ would tend towards if it had an infinite capacity and was trained with an infinite dataset verifying the brightness consistency assumption and without any uniformly textured surface. In this situation, the model is always capable of finding the unique depth prediction that minimizes the reconstruction error. In our experiments, we assume that the datasets are sufficiently large, that their images respect the aforementioned ideal properties and that the neural network architecture is wide enough to apply the approximation in \cref{eq:approx}.

Finally, it is sufficient to have only the marginal values of the multivariate distribution $D$ to compute the expectation in \cref{eq:approx} since the $\text{recons}(.,.)$ function only applies pixel-wise operations (see proof in the supplementary material).

\paragraph{Sampling strategy.}

The expectation in \cref{eq:approx} is approximated with a sampling strategy. The samples cannot be random because this would not allow computing the gradients of the distribution parameters that are later used in the backpropagation algorithm to update the parameters $\theta$ of $\psi_{\theta}$.
Instead, for each marginal distribution of $D$, a set of $n$ samples $\mathcal{S}_{\eta}$ is defined so as to accurately represent the predicted distribution (more details in the supplementary material). At the end of the sampling stage, $n$ depth maps of size $HW$ are obtained, from which $n$ reconstructions are computed. These reconstructions are weighted according to the sample used and summed.
% Instead, a set of $n$ samples $\mathcal{S}_{\eta}$ is pre-defined according to the family of distribution of $D$ and then adapted for each pixel depending on the predicted parameters of the distribution so as samples accurately represent the predicted distribution $D$ (more details in supplementary material). This allows the sampling strategy to be representative of the predicted distribution of $D$. For each sample, a reconstruction image is computed. 
Thus, the final reconstruction loss $\mathcal{L}_{\text{recons}}$ is:
\begin{equation}
    \mathcal{L}_{\text{recons}}(\theta) = \text{err}\left( \frac{\sum_{d \in \mathcal{S}_{\eta}} \text{p}_D(d|\eta) \text{recons}(d, I_s^*)}{\sum_{d \in \mathcal{S}_{\eta}} \text{p}_D(d|\eta)} , I_t^* \right).
\end{equation}

\begin{table*}[]
\scriptsize
\centering
\begin{tabular}{|c|c|c|c|c|c|cc|cc|cc|c|c|}
\hline
 &  &  & \cellcolor[HTML]{E8A194} & \cellcolor[HTML]{E8A194} & \cellcolor[HTML]{94BBE8} & \multicolumn{2}{c|}{Abs Rel} & \multicolumn{2}{c|}{RMSE} & \multicolumn{2}{c|}{$\delta < 1.25$} & \cellcolor[HTML]{E8A194} & \cellcolor[HTML]{E8A194} \\ \cline{7-12}
\multirow{-2}{*}{Sup} & \multirow{-2}{*}{Resolution} & \multirow{-2}{*}{\begin{tabular}[c]{@{}c@{}}Self-\\ dist\end{tabular}} & \multirow{-2}{*}{\cellcolor[HTML]{E8A194}Abs Rel $\downarrow$} & \multirow{-2}{*}{\cellcolor[HTML]{E8A194}RMSE $\downarrow$} & \multirow{-2}{*}{\cellcolor[HTML]{94BBE8}$\delta < 1.25$ $\uparrow$} & \multicolumn{1}{c|}{\cellcolor[HTML]{E8A194}AUSE $\downarrow$} & \cellcolor[HTML]{94BBE8}AURG$\uparrow$ & \multicolumn{1}{c|}{\cellcolor[HTML]{E8A194}AUSE $\downarrow$} & \cellcolor[HTML]{94BBE8}AURG$\uparrow$ & \multicolumn{1}{c|}{\cellcolor[HTML]{E8A194}AUSE $\downarrow$} & \cellcolor[HTML]{94BBE8}AURG$\uparrow$ & \multirow{-2}{*}{\cellcolor[HTML]{E8A194}ARU$\downarrow$} & \multirow{-2}{*}{\cellcolor[HTML]{E8A194}RMSU$\downarrow$} \\ \hline
M & $640 \times 192$ &  & 0.084 & 3.621 & 0.920 & \multicolumn{1}{c|}{0.025} & 0.030 & \multicolumn{1}{c|}{0.744} & 2.365 & \multicolumn{1}{c|}{0.025} & 0.048 & 0.074 & 3.405 \\
M & $640 \times 192$ & \checkmark & \textbf{0.082} & {\ul 3.570} & {\ul 0.926} & \multicolumn{1}{c|}{{\ul 0.022}} & 0.031 & \multicolumn{1}{c|}{\textbf{0.315}} & \textbf{2.728} & \multicolumn{1}{c|}{{\ul 0.015}} & {\ul 0.054} & {\ul 0.063} & 2.870 \\
M & $1024 \times 320$ &  & 0.087 & 3.653 & 0.922 & \multicolumn{1}{c|}{0.028} & 0.030 & \multicolumn{1}{c|}{0.543} & 2.574 & \multicolumn{1}{c|}{0.022} & 0.052 & 0.064 & \textbf{2.720} \\
M & $1024 \times 320$ & \checkmark & {\ul 0.083} & \textbf{3.481} & \textbf{0.928} & \multicolumn{1}{c|}{\textbf{0.020}} & \textbf{0.034} & \multicolumn{1}{c|}{{\ul 0.324}} & {\ul 2.640} & \multicolumn{1}{c|}{\textbf{0.011}} & \textbf{0.056} & \textbf{0.061} & {\ul 2.751} \\ \hline
S & $640 \times 192$ &  & 0.080 & 3.653 & 0.920 & \multicolumn{1}{c|}{0.023} & 0.030 & \multicolumn{1}{c|}{0.687} & 2.467 & \multicolumn{1}{c|}{0.022} & {\ul 0.051} & 0.072 & 3.404 \\
S & $640 \times 192$ & \checkmark & 0.079 & 3.606 & 0.922 & \multicolumn{1}{c|}{\textbf{0.015}} & \textbf{0.037} & \multicolumn{1}{c|}{{\ul 0.303}} & \textbf{2.811} & \multicolumn{1}{c|}{{\ul 0.013}} & \textbf{0.058} & 0.075 & 3.451 \\
S & $1024 \times 320$ &  & {\ul 0.074} & {\ul 3.361} & {\ul 0.931} & \multicolumn{1}{c|}{{\ul 0.022}} & 0.028 & \multicolumn{1}{c|}{0.597} & 2.304 & \multicolumn{1}{c|}{0.021} & 0.043 & \textbf{0.065} & \textbf{3.061} \\
S & $1024 \times 320$ & \checkmark & \textbf{0.073} & \textbf{3.315} & \textbf{0.933} & \multicolumn{1}{c|}{\textbf{0.015}} & {\ul 0.034} & \multicolumn{1}{c|}{\textbf{0.277}} & {\ul 2.584} & \multicolumn{1}{c|}{\textbf{0.012}} & {\ul 0.051} & {\ul 0.069} & {\ul 3.172} \\ \hline
MS & $640 \times 192$ &  & 0.080 & 3.486 & 0.924 & \multicolumn{1}{c|}{\textbf{0.015}} & \textbf{0.038} & \multicolumn{1}{c|}{{\ul 0.288}} & \textbf{2.701} & \multicolumn{1}{c|}{\textbf{0.013}} & \textbf{0.058} & 0.076 & 3.336 \\
MS & $640 \times 192$ & \checkmark & 0.079 & 3.456 & 0.925 & \multicolumn{1}{c|}{0.025} & 0.028 & \multicolumn{1}{c|}{0.602} & 2.358 & \multicolumn{1}{c|}{0.024} & 0.045 & {\ul 0.071} & 3.175 \\
MS & $1024 \times 320$ &  & {\ul 0.076} & {\ul 3.338} & {\ul 0.927} & \multicolumn{1}{c|}{0.023} & 0.027 & \multicolumn{1}{c|}{0.571} & 2.296 & \multicolumn{1}{c|}{{\ul 0.022}} & 0.046 & \textbf{0.067} & \textbf{3.057} \\
MS & $1024 \times 320$ & \checkmark & \textbf{0.075} & \textbf{3.241} & \textbf{0.929} & \multicolumn{1}{c|}{{\ul 0.016}} & {\ul 0.033} & \multicolumn{1}{c|}{\textbf{0.275}} & {\ul 2.499} & \multicolumn{1}{c|}{\textbf{0.013}} & {\ul 0.052} & {\ul 0.071} & {\ul 3.109} \\ \hline
\end{tabular}
\caption{Results of the M, S and MS MonoProb methods with the Resnet50 architecture and two different resolutions with and without self-distillation. These demonstrate the ability of our MonoProb method to work with other architectures and high-resolution images.}
\label{tab:res_hr_resnet50}
\end{table*}

\begin{table*}[]
\scriptsize
\centering
\begin{tabular}{|c|c|c|c|c|cc|cc|cc|c|c|}
\hline
 &  & \cellcolor[HTML]{E8A194} & \cellcolor[HTML]{E8A194} & \cellcolor[HTML]{94BBE8} & \multicolumn{2}{c|}{Abs Rel} & \multicolumn{2}{c|}{RMSE} & \multicolumn{2}{c|}{$\delta < 1.25$} & \cellcolor[HTML]{E8A194} & \cellcolor[HTML]{E8A194} \\
\multirow{-2}{*}{Dataset} & \multirow{-2}{*}{Method} & \multirow{-2}{*}{\cellcolor[HTML]{E8A194}Abs Rel $\downarrow$} & \multirow{-2}{*}{\cellcolor[HTML]{E8A194}RMSE $\downarrow$} & \multirow{-2}{*}{\cellcolor[HTML]{94BBE8}$\delta < 1.25$ $\uparrow$} & \multicolumn{1}{c|}{\cellcolor[HTML]{E8A194}AUSE $\downarrow$} & \cellcolor[HTML]{94BBE8}AURG$\uparrow$ & \multicolumn{1}{c|}{\cellcolor[HTML]{E8A194}AUSE $\downarrow$} & \cellcolor[HTML]{94BBE8}AURG$\uparrow$ & \multicolumn{1}{c|}{\cellcolor[HTML]{E8A194}AUSE $\downarrow$} & \cellcolor[HTML]{94BBE8}AURG$\uparrow$ & \multirow{-2}{*}{\cellcolor[HTML]{E8A194}ARU$\downarrow$} & \multirow{-2}{*}{\cellcolor[HTML]{E8A194}RMSU$\downarrow$} \\ \hline
 & \cite{godard2019digging} & \textbf{0.322} & 7.417 & - & \multicolumn{1}{c|}{-} & - & \multicolumn{1}{c|}{-} & - & \multicolumn{1}{c|}{-} & - & - & - \\
 & \cite{poggi2020uncertainty}-Self & 0.334 & 6.840 & {\ul 0.514} & \multicolumn{1}{c|}{0.173} & 0.029 & \multicolumn{1}{c|}{4.954} & 0.065 & \multicolumn{1}{c|}{0.251} & 0.036 & 0.286 & 6.500 \\
 & \textbf{Ours} & 0.333 & {\ul 6.729} & {\ul 0.514} & \multicolumn{1}{c|}{{\ul 0.124}} & {\ul 0.079} & \multicolumn{1}{c|}{{\ul 1.966}} & {\ul 2.977} & \multicolumn{1}{c|}{{\ul 0.231}} & {\ul 0.060} & {\ul 0.271} & {\ul 5.705} \\
\multirow{-4}{*}{Make3D} & \textbf{Ours-self} & {\ul 0.327} & \textbf{6.687} & \textbf{0.521} & \multicolumn{1}{c|}{\textbf{0.112}} & \textbf{0.087} & \multicolumn{1}{c|}{\textbf{1.583}} & \textbf{3.335} & \multicolumn{1}{c|}{\textbf{0.219}} & \textbf{0.070} & \textbf{0.267} & \textbf{5.781} \\ \hline
 & \cite{godard2019digging} & 0.226 & 10.043 & 0.653 & \multicolumn{1}{c|}{-} & - & \multicolumn{1}{c|}{-} & - & \multicolumn{1}{c|}{-} & -  & -  & - \\
 & \cite{poggi2020uncertainty}-Self & 0.220 & 9.765 & 0.662 & \multicolumn{1}{c|}{0.127} & 0.009 & \multicolumn{1}{c|}{7.446} & 0.575 & \multicolumn{1}{c|}{0.224} & 0.017 & 0.197 & 9.518 \\
 & \textbf{Ours} & {\ul 0.224} & {\ul 9.757} & {\ul 0.666} & \multicolumn{1}{c|}{{\ul 0.081}} & {\ul 0.053} & \multicolumn{1}{c|}{{\ul 3.148}} & {\ul 4.905} & \multicolumn{1}{c|}{{\ul 0.128}} & {\ul 0.118} & {\ul 0.162} & {\ul 8.165} \\
\multirow{-4}{*}{Nuscenes} & \textbf{Ours-self} & \textbf{0.219} & \textbf{9.559} & \textbf{0.670} & \multicolumn{1}{c|}{\textbf{0.0720}} & \textbf{0.060} & \multicolumn{1}{c|}{\textbf{1.644}} & \textbf{6.218} & \multicolumn{1}{c|}{\textbf{0.099}} & \textbf{0.141} & \textbf{0.175} & \textbf{8.498} \\ \hline
\end{tabular}
\caption{Evaluation on the Make3D and Nuscenes datasets of our M models trained on KITTI with monocular images only shows that our method generalizes well to other datasets both for the depth and the uncertainty.}
\label{tab:general}
\end{table*}

\subsection{Self-distillation training}

\cite{poggi2020uncertainty} is the first unsupervised monocular depth estimation method that requires only one inference to provide an interpretable uncertainty. It consists in training an unsupervised monocular depth estimation \textit{teacher} model without uncertainty, then using its predictions as pseudo ground truth in the supervised training of a  \textit{student} model. This \textit{student} model returns the parameters of a probability distribution of the depth $D_s$. The loss function is the negative log-likelihood. We redesign this loss in case the \textit{teacher} already returns a distribution $D_t$. We train the \textit{student} so that its predicted distribution matches the frozen \textit{teacher} distribution. For this purpose, our loss function is the Kullback-Leibler divergence between the \textit{teacher} and the \textit{student} distributions. Following the results in \cref{tab:abl_dist}, we choose the Gaussian distribution as the family of distributions for the \textit{teacher}: $D_t = \mathcal{N}(\mu_t, \sigma_t)$ and the \textit{student}: $D_s = \mathcal{N}(\mu_s, \sigma_s)$. Thus, the self-distillation loss is $\mathcal{L}_{\text{self}}$:
\begin{equation}
    \mathcal{L}_{\text{self}}(\theta) = D_{KL}(D_s\|D_t) = \log\frac{\sigma_t}{\sigma_s} + \frac{\sigma_s^2+(\mu_s-\mu_t)^2}{2\sigma_t^2}.
\end{equation}

\section{Experiments}

\begin{table*}[]
\scriptsize
\centering
\begin{tabular}{|c|c|c|c|cc|cc|cc|c|c|}
\hline
 & \cellcolor[HTML]{E8A194} & \cellcolor[HTML]{E8A194} & \cellcolor[HTML]{94BBE8} & \multicolumn{2}{c|}{Abs Rel} & \multicolumn{2}{c|}{RMSE} & \multicolumn{2}{c|}{$\delta < 1.25$} & \cellcolor[HTML]{E8A194} & \cellcolor[HTML]{E8A194} \\ \cline{5-10}
\multirow{-2}{*}{\begin{tabular}[c]{@{}c@{}}Uncertainty\\ prediction\end{tabular}} & \multirow{-2}{*}{\cellcolor[HTML]{E8A194}Abs Rel $\downarrow$} & \multirow{-2}{*}{\cellcolor[HTML]{E8A194}RMSE $\downarrow$} & \multirow{-2}{*}{\cellcolor[HTML]{94BBE8}$\delta < 1.25$ $\uparrow$} & \multicolumn{1}{c|}{\cellcolor[HTML]{E8A194}AUSE $\downarrow$} & \cellcolor[HTML]{94BBE8}AURG$\uparrow$ & \multicolumn{1}{c|}{\cellcolor[HTML]{E8A194}AUSE $\downarrow$} & \cellcolor[HTML]{94BBE8}AURG$\uparrow$ & \multicolumn{1}{c|}{\cellcolor[HTML]{E8A194}AUSE $\downarrow$} & \cellcolor[HTML]{94BBE8}AURG$\uparrow$ & \multirow{-2}{*}{\cellcolor[HTML]{E8A194}ARU$\downarrow$} & \multirow{-2}{*}{\cellcolor[HTML]{E8A194}RMSU$\downarrow$} \\ \hline
$\sigma$ & 0.418 & 12.047 & 0.321 & \multicolumn{1}{c|}{0.226} & 0.003 & \multicolumn{1}{c|}{9.087} & 0.045 & \multicolumn{1}{c|}{0.346} & 0.006 & 0.418 & 12.047 \\
$\alpha$ & \textbf{0.089} & \textbf{3.852} & \textbf{0.914} & \multicolumn{1}{c|}{\textbf{0.031}} & \textbf{0.026} & \multicolumn{1}{c|}{\textbf{0.719}} & \textbf{2.560} & \multicolumn{1}{c|}{\textbf{0.030}} & \textbf{0.050} & \textbf{0.064} & \textbf{2.912} \\ \hline
\end{tabular}
\caption{The ablation on how to predict uncertainty in unsupervised training with probabilistic image reconstruction shows that directly predicting $\sigma$ is unable to train the model contrary to predicting $\alpha$ and then deducing $\sigma = \alpha\times\mu$.}
\label{tab:abl_alpha_mu}
\end{table*}

\begin{table*}[]
\scriptsize
\centering
\begin{tabular}{|c|c|c|c|cc|cc|cc|c|c|}
\hline
 & \cellcolor[HTML]{E8A194} & \cellcolor[HTML]{E8A194} & \cellcolor[HTML]{94BBE8} & \multicolumn{2}{c|}{Abs Rel} & \multicolumn{2}{c|}{RMSE} & \multicolumn{2}{c|}{$\delta < 1.25$} & \cellcolor[HTML]{E8A194} & \cellcolor[HTML]{E8A194} \\
\multirow{-2}{*}{\begin{tabular}[c]{@{}c@{}}Num\\ samples\end{tabular}} & \multirow{-2}{*}{\cellcolor[HTML]{E8A194}Abs Rel $\downarrow$} & \multirow{-2}{*}{\cellcolor[HTML]{E8A194}RMSE $\downarrow$} & \multirow{-2}{*}{\cellcolor[HTML]{94BBE8}$\delta < 1.25$ $\uparrow$} & \multicolumn{1}{c|}{\cellcolor[HTML]{E8A194}AUSE $\downarrow$} & \cellcolor[HTML]{94BBE8}AURG$\uparrow$ & \multicolumn{1}{c|}{\cellcolor[HTML]{E8A194}AUSE $\downarrow$} & \cellcolor[HTML]{94BBE8}AURG$\uparrow$ & \multicolumn{1}{c|}{\cellcolor[HTML]{E8A194}AUSE $\downarrow$} & \cellcolor[HTML]{94BBE8}AURG$\uparrow$ & \multirow{-2}{*}{\cellcolor[HTML]{E8A194}ARU$\downarrow$} & \multirow{-2}{*}{\cellcolor[HTML]{E8A194}RMSU$\downarrow$} \\ \hline
5 & 0.092 & 3.895 & 0.911 & \multicolumn{1}{c|}{0.033} & 0.025 & \multicolumn{1}{c|}{0.729} & \textbf{2.581} & \multicolumn{1}{c|}{0.033} & 0.049 & 0.066 & 2.968 \\
9 & \textbf{0.089} & \textbf{3.852} & \textbf{0.914} & \multicolumn{1}{c|}{\textbf{0.031}} & 0.026 & \multicolumn{1}{c|}{\textbf{0.719}} & 2.560 & \multicolumn{1}{c|}{\textbf{0.030}} & 0.050 & \textbf{0.064} & \textbf{2.912} \\
13 & 0.092 & 3.907 & 0.912 & \multicolumn{1}{c|}{0.032} & \textbf{0.027} & \multicolumn{1}{c|}{0.753} & 2.573 & \multicolumn{1}{c|}{\textbf{0.030}} & \textbf{0.051} & 0.066 & 2.952 \\ \hline
\end{tabular}
\caption{The ablation study on the number of samples provides an optimal value of 9 samples.}
\label{tab:abl_samples}
\end{table*}

\begin{table*}[]
\scriptsize
\centering
\begin{tabular}{|c|c|c|c|cc|cc|cc|c|c|}
\hline
 & \cellcolor[HTML]{E8A194} & \cellcolor[HTML]{E8A194} & \cellcolor[HTML]{94BBE8} & \multicolumn{2}{c|}{Abs Rel} & \multicolumn{2}{c|}{RMSE} & \multicolumn{2}{c|}{$\delta < 1.25$} & \cellcolor[HTML]{E8A194} & \cellcolor[HTML]{E8A194} \\
\multirow{-2}{*}{\begin{tabular}[c]{@{}c@{}}Family of\\ distributions\end{tabular}} & \multirow{-2}{*}{\cellcolor[HTML]{E8A194}Abs Rel $\downarrow$} & \multirow{-2}{*}{\cellcolor[HTML]{E8A194}RMSE $\downarrow$} & \multirow{-2}{*}{\cellcolor[HTML]{94BBE8}$\delta < 1.25$ $\uparrow$} & \multicolumn{1}{c|}{\cellcolor[HTML]{E8A194}AUSE $\downarrow$} & \cellcolor[HTML]{94BBE8}AURG$\uparrow$ & \multicolumn{1}{c|}{\cellcolor[HTML]{E8A194}AUSE $\downarrow$} & \cellcolor[HTML]{94BBE8}AURG$\uparrow$ & \multicolumn{1}{c|}{\cellcolor[HTML]{E8A194}AUSE $\downarrow$} & \cellcolor[HTML]{94BBE8}AURG$\uparrow$ & \multirow{-2}{*}{\cellcolor[HTML]{E8A194}ARU$\downarrow$} & \multirow{-2}{*}{\cellcolor[HTML]{E8A194}RMSU$\downarrow$} \\ \hline
Laplace & 0.091 & 3.913 & \textbf{0.914} & \multicolumn{1}{c|}{0.032} & \textbf{0.027} & \multicolumn{1}{c|}{0.857} & 2.481 & \multicolumn{1}{c|}{\textbf{0.030}} & \textbf{0.050} & 0.099 & 3.816 \\
Normal & \textbf{0.089} & \textbf{3.852} & \textbf{0.914} & \multicolumn{1}{c|}{\textbf{0.031}} & 0.026 & \multicolumn{1}{c|}{\textbf{0.719}} & \textbf{2.560} & \multicolumn{1}{c|}{\textbf{0.030}} & \textbf{0.050} & \textbf{0.064} & \textbf{2.912} \\ \hline
\end{tabular}
\caption{The ablation of the family of distributions for $D$ shows that the normal distribution performs better than the Laplace distribution.}
\label{tab:abl_dist}
\end{table*}

\begin{table*}[h]
\scriptsize
\centering
\begin{tabular}{|c|c|c|c|cc|cc|cc|c|c|}
\hline
 & \cellcolor[HTML]{E8A194} & \cellcolor[HTML]{E8A194} & \cellcolor[HTML]{94BBE8} & \multicolumn{2}{c|}{Abs Rel} & \multicolumn{2}{c|}{RMSE} & \multicolumn{2}{c|}{$\delta < 1.25$} & \cellcolor[HTML]{E8A194} & \cellcolor[HTML]{E8A194} \\ \cline{5-10}
\multirow{-2}{*}{Methods} & \multirow{-2}{*}{\cellcolor[HTML]{E8A194}Abs Rel$\downarrow$} & \multirow{-2}{*}{\cellcolor[HTML]{E8A194}RMSE$\downarrow$} & \multirow{-2}{*}{\cellcolor[HTML]{94BBE8}$\delta < 1.25\uparrow$} & \multicolumn{1}{c|}{\cellcolor[HTML]{E8A194}AUSE$\downarrow$} & \cellcolor[HTML]{94BBE8}AURG$\uparrow$ & \multicolumn{1}{c|}{\cellcolor[HTML]{E8A194}AUSE$\downarrow$} & \cellcolor[HTML]{94BBE8}AURG$\uparrow$ & \multicolumn{1}{c|}{\cellcolor[HTML]{E8A194}AUSE$\downarrow$} & \cellcolor[HTML]{94BBE8}AURG$\uparrow$ & \multirow{-2}{*}{\cellcolor[HTML]{E8A194}ARU$\downarrow$} & \multirow{-2}{*}{\cellcolor[HTML]{E8A194}RMSU$\downarrow$} \\ \hline
NLL & 0.088 & 3.785 & 0.917 & \multicolumn{1}{c|}{\textbf{0.022}} & 0.034 & \multicolumn{1}{c|}{0.330} & 2.897 & \multicolumn{1}{c|}{0.016} & \textbf{0.061} & \textbf{0.066} & \textbf{2.992} \\
KL-Div & \textbf{0.087} & \textbf{3.781} & \textbf{0.919} & \multicolumn{1}{c|}{\textbf{0.022}} & \textbf{0.035} & \multicolumn{1}{c|}{\textbf{0.322}} & \textbf{2.904} & \multicolumn{1}{c|}{\textbf{0.015}} & \textbf{0.060} & \textbf{0.066} & 2.997 \\ \hline
\end{tabular}
\caption{Comparing the performance of self-distillation with negative log-likelihood (NLL) where the pseudo ground truth is a depth map, and the Kullback-Leibler divergence (KL-Div) where the pseudo ground truth is a map of depth distributions, highlights a slight improvement with the Kullback-Leibler divergence.}
\label{tab:abl_self_dist}
\end{table*}

\subsection{Absolute uncertainty metrics}\label{sec:metrics}

The depth metrics are those of \cite{poggi2020uncertainty}, in particular the absolute relative error (Abs Rel), the root mean square error (RMSE), and the amount of inliners ($\delta<1.25$). We use the STD of the predicted distributions as a pixel-wise uncertainty estimate. Following \cite{poggi2020uncertainty}, we compute the Area Under the Sparsification Error (AUSE) and the Area Under the Random Gain (AURG), which are relative metrics within an image. They measure the ability of the uncertainty estimation to sort pixels in order of descending error $\epsilon$ for a given depth error metric in an image. AUSE compares the sparsification curve obtained by this sorting to the perfect sparsification curve obtained by sorting pixels directly based on their error. AURG quantifies the improvement of the sparsification curve of the predicted uncertainty relative to a random sparsification curve where no uncertainty modeling is performed. More details about these metrics can be found in \cite{poggi2020uncertainty}.

These metrics are designed to compare any kind of uncertainty metrics even if they are not directly related to depth but rather to the reconstruction quality for instance, as in \cite{klodt2018supervising, poggi2020uncertainty}. However, they suffer from some drawbacks: they cannot provide an absolute measure of uncertainty and they are not consistent from one image to another. Indeed, since they are based on sorting uncertainties within an image, these metrics are invariant within a growing bijection on the predicted uncertainties. Likewise, a given pair of uncertainty and corresponding error, will not have the same contribution to the global uncertainty metric depending on the performance of the other pixels of the same image. 

Therefore, we introduce two new uncertainty metrics that measure the ability to anticipate the true depth error. Thus, we propose the Absolute Relative Uncertainty (ARU), which is relative to the ground truth depth, and the Root Mean Square Uncertainty Error (RMSU), which is not. Let $D^*$, $\hat{D}$ and $U$ be respectively the ground truth depth map, the predicted depth map and the predicted uncertainty map for a single image:

\begin{equation}
    \begin{split}
        \text{ARU} & = \|(U - |\hat{D} - D^*|) \oslash D^*\|_1/HW \\
        \text{RMSU} & =  \sqrt{\|(U - |\hat{D} - D^*|)^2\|_1/HW},
    \end{split}
\end{equation}

\noindent where $|.|$, $\oslash$ and $\|.\|_1$ denote the element-wise absolute value, the element-wise division and the $\ell_1$-norm, respectively. These metrics are then averaged over all the ground truth depth maps of the dataset. By construction, these metrics are only suitable for quantifying uncertainty that is directly related to the depth estimation error.

\subsection{Implementation details}

We implement our method on top of Monodepth2 \cite{godard2019digging}, an unsupervised monocular depth estimation approach that is trained with either monocular videos only (M), pairs of stereo image pairs only (S), or both monocular videos and stereo images (MS). When monocular videos are used for training (M or MS), an additional \textit{pose} network is jointly trained with the depth network to provide the camera motion $T$. In contrast, when stereo images are used the camera motion is given by the translation between the two stereo cameras. Monodepth2 is also designed to handle occlusions and ignore dynamic objects that would disrupt the training. Thus, following \cite{godard2019digging}, we finetune a Resnet18 network \cite{he2016deep} pretrained on the ImageNet dataset \cite{deng2009imagenet} for 20 epochs with the Adam optimizer \cite{kingma2014adam}. We use batches of 12 images resized to $192 \times 640$ and augmented as in \cite{godard2019digging}. We choose a Gaussian distribution $\mathcal{N}(\mu, \sigma)$ as the family of distributions for depth, so our depth network must return a two-channel map, one for each parameter. On one of them, we apply a sigmoid activation function to predict a disparity from which the mean $\mu$ of the depth distribution is deduced following \cite{godard2019digging}. When training with our probabilistic reconstruction loss, a sigmoid is also applied to the second channel so that it outputs a scalar $\alpha\in [0, 1]$ that weights the mean depth $\mu$ to obtain the STD $\sigma = \alpha \times \mu$. Instead, at the self-distillation stage, the network directly returns the $\sigma$ by applying an exponential activation function as in \cite{poggi2020uncertainty}. The number of samples is set to 9. The learning rate is set to $10^{-4}$ and dropped to $10^{-5}$ over the last 5 epochs. We carry out experiments on the KITTI dataset \cite{kitti}, which is composed of 42K images extracted from 61 driving videos. Following common practice, we employ the popular Eigen split \cite{eigen2014depth}. At evaluation, we cap depth to 80 meters and use the improved ground truth provided by \cite{uhrig2017sparsity}. The uncertainty is the STD of the predicted distributions.

\subsection{Results on KITTI}

We conducted experiments on the three depth estimation paradigms of \cite{godard2019digging}: M, S and MS. In \cref{tab:res}, we compare the performance obtained with our baseline \cite{godard2019digging} that does not provide uncertainty and with predictive methods implemented in \cite{poggi2020uncertainty} (i.e. methods that directly predict an uncertainty quantification within a single inference). In accordance with \cref{sec:metrics}, we cannot compute our new metrics (ARU and RMSU) on \cite{poggi2020uncertainty}-Repr and \cite{poggi2020uncertainty}-Log. Indeed, these methods respectively provide as uncertainty quantification, an estimate of the reprojection error and an estimate of the scale parameter of the distribution of the reconstruction error. For fair comparisons, we report the number of trainings (\#Trn). In the following, we refer to our MonoProb method without self-distillation as raw-MonoProb and to our MonoProb with self-distillation as self-distilled-MonoProb. Qualitative results in \cref{fig:visu} show that \cite{poggi2020uncertainty}-Self tends to underestimate the uncertainty of distant objects and background, and to overestimate the uncertainty of object edges compared to our methods (more qualitative results on KITTI including depth maps are available in the supplementary material).

First, let's describe the results of single training methods. For the M training, raw-MonoProb performs slightly better on depth and significantly improves performance on uncertainty metrics. For the S training, the same behavior is observed for depth metrics: raw-MonoProb is a bit more efficient on depth than the other methods that require a single training. In terms of uncertainty, our raw-MonoProb results tend to be equivalent to \cite{poggi2020uncertainty}-Log. For the MS training, the depth performance of our raw-MonoProb is slightly below \cite{godard2019digging} and very close to other methods that require a single training to predict an uncertainty. The results on uncertainty metrics show that our method is either better than or very close to other comparable methods. For self-distilled methods, we observe the same behavior regardless of the training paradigm (M, S, or MS). Our self-distilled-MonoProb performs similarly to \cite{poggi2020uncertainty}-Self (trained with self-distillation on scalar pseudo ground truth) on depth metrics but the uncertainties of self-distilled-MonoProb are significantly better than those of\cite{poggi2020uncertainty}-Self. Interestingly we show that our raw-MonoProb outperforms \cite{poggi2020uncertainty}-Self on our new metrics ARU and RMSU. We also show in the supplementary material that our MonoProb has similar or better performance than uncertainty methods that require more than one inference.

We evaluate our M model on KITTI with the raw ground truth for comparison with the VDN approach \cite{dikov2022variational} in \cref{tab:res_VDN}. This shows a higher effectiveness of our method for both depth estimation and uncertainty quantification. Finally, we provide results of our method with the Resnet50  architecture \cite{he2016deep} and high-resolution images in \cref{tab:res_hr_resnet50}. This shows that MonoProb also works on different architectures and image resolutions as the results are better than those of \cref{tab:res}.

\subsection{Generalization to other datasets}

We also evaluate the generalization ability of our method on two other datasets of outdoor and urban scenes: Make3D \cite{make3d} and Nuscenes \cite{nuscenes}. The results on both datasets (see \cref{tab:general} and supplementary for qualitative results) show that our method without self-distillation is equivalent to the self-distilled approach of \cite{poggi2020uncertainty} in terms of depth performance and better at estimating uncertainty. Our self-distilled models are more accurate than other methods in both depth and uncertainty estimation. This shows that our method has a good generalization ability on images from different datasets.

\subsection{Ablation}

To justify the impact of our contributions and the interest of our implementation choices, we conduct an ablation study including the computation of the STD $\sigma$, the family distributions for depth, the number of samples and the self-distillation loss. First, we show the benefit of predicting $\alpha$ before deducing $\sigma=\alpha\times\mu$ instead of directly predicting $\sigma$ when training with our probabilistic reconstruction loss. Experiments in \cref{tab:abl_alpha_mu} highlight that a direct prediction of $\sigma$ prevents the network from learning anything, leading to very low performance compared to our strategy of considering the $\sigma$ as a fraction of the mean depth $\mu$. Studying the number of samples yields an optimal value of 9 in \cref{tab:abl_samples}. The ablation in \cref{tab:abl_dist} highlights that normal distributions perform better overall than Laplace distributions as the family of distributions for depth. Finally, the experiments in \cref{tab:abl_self_dist} show the influence of our self-distillation loss: fitting two distributions with the Kullback-Leibler divergence outperforms minimizing the negative log-likelihood. As a matter of reproducibility, we provide the expression of the negative log-likelihood loss $\mathcal{L}_{\text{NLL}}$ we use in this ablation. For a fair comparison, we assume the student follows a normal distribution $D_s = \mathcal{N}(\mu_s, \sigma_s)$:
\begin{equation}
    \mathcal{L}_{\text{NLL}}(\theta) = (\mu_s - \mu_t)^2/(2\sigma_s^2) + \log(\sigma_s),
\end{equation}

\noindent $\mu_t$ being the mean of the teacher's depth distribution used as pseudo ground truth.

\section{Conclusion}

In this paper, we propose MonoProb an unsupervised method for training monocular depth estimation networks that returns an interpretable uncertainty within a single inference. This uncertainty, which anticipates the prediction errors, is the STD of the depth distribution returned by the depth estimation network. We also introduce a new self-distilled loss to train another network using a depth distribution as pseudo ground truth. Finally, we design new metrics that are better suited to measure the performance of interpretable uncertainty, i.e., uncertainty that is a direct anticipation of depth prediction errors. Through extensive experiments, we highlight that MonoProb improves performance relative to other unsupervised depth estimation methods that provide a quantification of uncertainty. We also demonstrate the cross-domain generalization ability of our method, that it works on different neural network architectures and on high-resolution images.

\paragraph{Acknowledgment} This work was performed using HPC resources from GENCI-IDRIS (Grant 2023-AD011013778) and the FactoryIA supercomputer, financially supported by the Ile-De-France Regional Council. Thanks to Riccardo Finotello for his helpful discussions.

\appendix
\appendixpage

% \section*{\Large \centering Appendix}

% \vspace{0.5cm}

\section{Proof that we only need the marginals of the depth distribution}

Let $N\in \mathbb{N}_+^*$, $D=[D_1, ..., D_{N}]$ be a multi-variate random variable and $\text{recons}(.,I_s^*)$ a function so that:
\begin{align*}
     \text{recons}(.,I_s^*)\colon & \mathbb{R}^{N} \longrightarrow \mathbb{R}^{N\times 3 } \\
     & d\mapsto[\text{recons}_1(d_1,I_s^*), ..., \text{recons}_{N}(d_{N},I_s^*)], \\
\end{align*}
where $\forall\;i \in \{1, ..., N\}$ and $x \in \mathbb{R}$, $\text{recons}_i(x,I_s^*) \in \mathbb{R}^{3}$ and $\eta \in \mathbb{R}^{N\times 2}$.
\begin{equation*}
    \begin{split}
        & \mathbb{E}_D[\text{recons}(D, I_s^*)|\eta] \\
        = & \int_{\mathcal{D}}[\text{recons}_1(d_1,I_s^*), ..., \text{recons}_{N}(d_{N},I_s^*)] p_D(d|\eta)\text{d}d \\
        = & \Bigg[\int_{\mathcal{D}}\text{recons}_1(d_1, I_s^*)p_D(d|\eta)\text{d}d, ..., \\ 
        & \hspace*{3cm}\int_{\mathcal{D}}\text{recons}_{N}(d_{N}, I_s^*)p_D(d|\eta)\text{d}d \Bigg] \\
        = & \Bigg[\int_{\mathcal{D}}\text{recons}_1(d_1, I_s^*)p_{D_1}(d_1|\eta)\text{d}d_1, ..., \\
        & \hspace*{1.45cm}\int_{\mathcal{D}}\text{recons}_{N}(d_{N}, I_s^*)p_{D_{N}}(d_{N}|\eta)\text{d}d_{N}\Bigg] \\
        = & [\mathbb{E}_{D_1}[\text{recons}_1(D_1, I_s^*)|\eta], ..., \\
        & \hspace*{3.5cm}\mathbb{E}_{D_{N}}[\text{recons}_{N}(D_{N}, I_s^*)|\eta]].
    \end{split}
\end{equation*}

This shows that the marginals of $D$ are sufficient to compute $\mathbb{E}_D[\text{recons}(D, I_s^*)|\eta]$.

\section{Sampling strategy}

The MonoProb depth estimator returns a map of $HW$ univariate depth distributions of parameters $\eta$ that are independently sampled $n$ times so that for each distribution $D_k$ with $k \in \llbracket 1, HW \rrbracket$, the sample set $\mathcal{S}_{\eta}^k$ accurately represents the corresponding distributions. Finally, $n$ depth maps of size $HW$ are obtained. We design our sampling strategy for families of depth distributions belonging to the symmetric generalized normal distribution with a density of the form:

\begin{equation}
    \begin{split}
        f: & \: \mathbb{R}\rightarrow \mathbb{R}\\
        & x\mapsto \frac{\beta}{2\gamma\Gamma(1/\beta)}\exp\left(-(|x-\mu|/\gamma)^{\beta}\right)
    \end{split}
\end{equation}

where $\mu$ is the mean, $\beta$ is the shape parameter (1 for a Laplace distribution and 2 for a Gaussian distribution) and $\gamma$ is the scale parameter, which can be expressed as a function of the standard deviation $\sigma$ (so that $\gamma = \sigma/\sqrt{2}$ for the Laplace distribution and $\gamma = \sqrt{2}\sigma$ for the Gaussian distribution). We sample only indices for which the ratio between their density and the density of the mean $f(\mu)$ is in $\{\frac{2i}{n+1}\}_{i=1}^{\lfloor\frac{n+1}{2}\rfloor}$. This results in a set $\mathcal{S}_{\eta}^k$ of $n$ samples, symmetrically and evenly distributed around the mean. These samples are also controlled to be close enough to the mean so that their density is not too close to zero and so that at most one sample only is equal to the mean. Furthermore, the sampling strategy was chosen because the relationship between the samples and the parameters of the distribution $D_k$ facilitates the computation of the backpropagation compared to if we had used quantiles of the distributions. These samples are:
\begin{equation}
   \begin{split}
        \forall n>0, \mathcal{S}_{\eta}^k & = \left\{s \: \Big| \: \frac{f(s)}{f(\mu)} = \frac{2i}{n+1} \right\}_{i=1}^{\lfloor\frac{n+1}{2}\rfloor} \\
        & = \left\{s \: \Big| \: \exp(-(|s-\mu|/\gamma)^{\beta} = \frac{2i}{n+1} \right\}_{i=1}^{\lfloor\frac{n+1}{2}\rfloor} \\
        & = \left\{\mu \pm \gamma\left(-\log\left(\frac{2i}{n+1}\right)\right)^{\frac{1}{\beta}}\right\}_{i=1}^{\lfloor\frac{n+1}{2}\rfloor}
    \end{split}
\end{equation}

In the paper, we carried out experiments with $n \in \{5, 9, 13\}$. For $n=5$, the corresponding sample set $\mathcal{S}_{\eta}^k$ is:

\begin{equation}
    \begin{split}
        \mathcal{S}_{\eta}^k = \Bigg\{\mu + \delta \: \Bigg| \:  \delta \in \Bigg\{ & -\gamma\left(-\log\frac{1}{3}\right)^{\frac{1}{\beta}}, \\
        & -\gamma\left(-\log\frac{2}{3}\right)^{\frac{1}{\beta}}, \\ 
        & 0, \\
        & \gamma\left(-\log\frac{2}{3}\right)^{\frac{1}{\beta}}, \\
        & \gamma\left(-\log\frac{1}{3}\right)^{\frac{1}{\beta}} \Bigg\} \Bigg\}.
    \end{split}
\end{equation}

\section{Comparison with \cite{godard2019digging, poggi2020uncertainty}'s methods}

In \cref{tab:all_comparisons}, we compare our MonoProb methods with and without self-distillation to all the methods introduced in \cite{godard2019digging, poggi2020uncertainty}, including those that require more than one inference to predict an uncertainty. We provide results for the three training paradigms that are: monocular video supervision only (M), stereo supervision only (S), and both monocular video and stereo supervision (MS). For each method, we report the number of trainings (\#Trn) and the number of inferences (\#Inf) required to generate the depth and uncertainty at test time. \#Inf must not be confused with \#Fwd used in \cite{poggi2020uncertainty}, which is the number of forwards required at test time to estimate the depth only. The results of the concurrent methods were taken from \cite{godard2019digging,poggi2020uncertainty}. Our new metrics have been computed using the checkpoints given by \cite{godard2019digging, poggi2020uncertainty} and on methods that provide an interpretable uncertainty as defined in the paper. The results show that our MonoProb and self-distilled MonoProb methods have similar depth performance to the other approaches. Likewise, our MonoProb without self-distillation provides competitive results in terms of uncertainty estimation. Our self-distilled MonoProb method shows overall better uncertainty estimation performance than other approaches. It even compares favorably with the methods that require more than one inference to predict uncertainty.

\begin{table*}[]
\scriptsize
\centering
\hspace*{-0.9\leftmargin}\begin{tabular}{|c|c|c|c|c|c|c|cc|cc|cc|c|c|}
\hline
 &  &  &  & \cellcolor[HTML]{E8A194} & \cellcolor[HTML]{E8A194} & \cellcolor[HTML]{94BBE8} & \multicolumn{2}{c|}{Abs Rel} & \multicolumn{2}{c|}{RMSE} & \multicolumn{2}{c|}{$\delta < 1.25$} & \cellcolor[HTML]{E8A194} & \cellcolor[HTML]{E8A194} \\
\multirow{-2}{*}{Sup} & \multirow{-2}{*}{Methods} & \multirow{-2}{*}{\#Trn} & \multirow{-2}{*}{\#Inf} & \multirow{-2}{*}{\cellcolor[HTML]{E8A194}Abs Rel $\downarrow$} & \multirow{-2}{*}{\cellcolor[HTML]{E8A194}RMSE $\downarrow$} & \multirow{-2}{*}{\cellcolor[HTML]{94BBE8}$\delta < 1.25$ $\uparrow$} & \multicolumn{1}{c|}{\cellcolor[HTML]{E8A194}AUSE $\downarrow$} & \cellcolor[HTML]{94BBE8}AURG$\uparrow$ & \multicolumn{1}{c|}{\cellcolor[HTML]{E8A194}AUSE $\downarrow$} & \cellcolor[HTML]{94BBE8}AURG$\uparrow$ & \multicolumn{1}{c|}{\cellcolor[HTML]{E8A194}AUSE $\downarrow$} & \cellcolor[HTML]{94BBE8}AURG$\uparrow$ & \multirow{-2}{*}{\cellcolor[HTML]{E8A194}ARU$\downarrow$} & \multirow{-2}{*}{\cellcolor[HTML]{E8A194}RMSU$\downarrow$} \\ \hline
M & \cite{godard2019digging} & 1 & 1 & 0.090 & 3.942 & 0.914 & \multicolumn{1}{c|}{-} & - & \multicolumn{1}{c|}{-} & - & \multicolumn{1}{c|}{-} & - & - & - \\
M & \cite{godard2019digging}-Post & 1 & 2 & {\ul 0.088} & 3.841 & 0.917 & \multicolumn{1}{c|}{0.044} & 0.012 & \multicolumn{1}{c|}{2.864} & 0.412 & \multicolumn{1}{c|}{0.056} & 0.022 & 1.670 & 17.23 \\
M & \cite{poggi2020uncertainty}-Repr & 1 & 1 & 0.092 & 3.936 & 0.912 & \multicolumn{1}{c|}{0.051} & 0.008 & \multicolumn{1}{c|}{2.972} & 0.381 & \multicolumn{1}{c|}{0.069} & 0.013 & - & - \\
M & \cite{poggi2020uncertainty}-Log & 1 & 1 & 0.091 & 4.052 & 0.910 & \multicolumn{1}{c|}{0.039} & 0.020 & \multicolumn{1}{c|}{2.562} & 0.916 & \multicolumn{1}{c|}{{\ul 0.044}} & {\ul 0.038} & - & - \\
M & \cite{poggi2020uncertainty}-Self & 2 & 1 & \textbf{0.087} & 3.826 & \textbf{0.920} & \multicolumn{1}{c|}{0.030} & 0.026 & \multicolumn{1}{c|}{2.009} & 1.266 & \multicolumn{1}{c|}{0.030} & 0.045 & 0.074 & 3.730 \\
M & \cite{poggi2020uncertainty}-Drop & 1 & 8 & {\ul 0.101} & 4.146 & 0.892 & \multicolumn{1}{c|}{0.065} & 0.000 & \multicolumn{1}{c|}{2.568} & 0.944 & \multicolumn{1}{c|}{0.097} & 0.002 & 3.041 & 33.90 \\
M & \cite{poggi2020uncertainty}-Boot & 8 & 8 & 0.092 & 3.821 & 0.911 & \multicolumn{1}{c|}{0.058} & 0.001 & \multicolumn{1}{c|}{3.982} & -0.743 & \multicolumn{1}{c|}{0.084} & -0.001 & 0.791 & 8.635 \\
M & \cite{poggi2020uncertainty}-Snap & 1 & 8 & 0.091 & 3.921 & 0.912 & \multicolumn{1}{c|}{0.059} & -0.001 & \multicolumn{1}{c|}{3.979} & -0.639 & \multicolumn{1}{c|}{0.083} & -0.002 & 0.361 & 3.956 \\
M & \cite{poggi2020uncertainty}-Boot+Log & 8 & 8 & 0.092 & 3.850 & 0.910 & \multicolumn{1}{c|}{0.038} & 0.021 & \multicolumn{1}{c|}{2.449} & 0.820 & \multicolumn{1}{c|}{0.046} & 0.037 & - & - \\
M & \cite{poggi2020uncertainty}-Boot+Self & 9 & 8 & {\ul 0.088} & {\ul 3.799} & 0.918 & \multicolumn{1}{c|}{{\ul 0.029}} & {\ul 0.028} & \multicolumn{1}{c|}{1.924} & 1.316 & \multicolumn{1}{c|}{0.028} & 0.049 & 0.333 & 3.609 \\
M & \cite{poggi2020uncertainty}-Snap+Log & 1 & 8 & 0.092 & 3.961 & 0.911 & \multicolumn{1}{c|}{0.038} & 0.022 & \multicolumn{1}{c|}{2.385} & 1.001 & \multicolumn{1}{c|}{0.043} & 0.039 & - & - \\
M & \cite{poggi2020uncertainty}-Snap+Self & 2 & 8 & {\ul 0.088} & 3.832 & {\ul 0.919} & \multicolumn{1}{c|}{0.031} & 0.026 & \multicolumn{1}{c|}{2.043} & 1.230 & \multicolumn{1}{c|}{0.030} & 0.045 & 0.233 & \textbf{2.554} \\
M & Ours & 1 & 1 & 0.089 & 3.852 & 0.914 & \multicolumn{1}{c|}{0.031} & 0.026 & \multicolumn{1}{c|}{{\ul 0.719}} & {\ul 2.560} & \multicolumn{1}{c|}{0.030} & 0.050 & \textbf{0.064} & {\ul 2.912} \\
M & Ours-self & 2 & 1 & \textbf{0.087} & \textbf{3.762} & {\ul 0.919} & \multicolumn{1}{c|}{\textbf{0.022}} & \textbf{0.034} & \multicolumn{1}{c|}{\textbf{0.326}} & \textbf{2.880} & \multicolumn{1}{c|}{\textbf{0.014}} & \textbf{0.061} & {\ul 0.066} & 2.969 \\ \hline
S & \cite{godard2019digging} & 1 & 1 & 0.085 & 3.942 & 0.912 & \multicolumn{1}{c|}{-} & - & \multicolumn{1}{c|}{-} & - & \multicolumn{1}{c|}{-} & - & - & - \\
S & \cite{godard2019digging}-Post & 1 & 2 & {\ul 0.084} & {\ul 3.777} & {\ul 0.915} & \multicolumn{1}{c|}{0.036} & 0.020 & \multicolumn{1}{c|}{2.523} & 0.736 & \multicolumn{1}{c|}{0.044} & 0.034 & 0.292 & {\ul 3.016} \\
S & \cite{poggi2020uncertainty}-Repr & 1 & 1 & 0.085 & 3.873 & 0.913 & \multicolumn{1}{c|}{0.040} & 0.017 & \multicolumn{1}{c|}{2.275} & 1.074 & \multicolumn{1}{c|}{0.050} & 0.030 & - & - \\
S & \cite{poggi2020uncertainty}-Log & 1 & 1 & 0.085 & 3.860 & {\ul 0.915} & \multicolumn{1}{c|}{0.022} & 0.036 & \multicolumn{1}{c|}{0.938} & 2.402 & \multicolumn{1}{c|}{\textbf{0.018}} & {\ul 0.061} & - & - \\
S & \cite{poggi2020uncertainty}-Self & 2 & 1 & {\ul 0.084} & 3.835 & {\ul 0.915} & \multicolumn{1}{c|}{0.022} & 0.035 & \multicolumn{1}{c|}{1.679} & 1.642 & \multicolumn{1}{c|}{0.022} & 0.056 & 0.083 & 3.686 \\
S & \cite{poggi2020uncertainty}-Drop & 1 & 8 & 0.129 & 4.908 & 0.819 & \multicolumn{1}{c|}{0.103} & -0.029 & \multicolumn{1}{c|}{6.163} & -2.169 & \multicolumn{1}{c|}{0.231} & -0.080 & 5.494 & 61.84 \\
S & \cite{poggi2020uncertainty}-Boot & 8 & 8 & 0.085 & \textbf{3.772} & 0.914 & \multicolumn{1}{c|}{0.028} & 0.029 & \multicolumn{1}{c|}{2.291} & 0.964 & \multicolumn{1}{c|}{0.031} & 0.048 & 0.496 & 5.211 \\
S & \cite{poggi2020uncertainty}-Snap & 1 & 8 & 0.085 & 3.849 & 0.912 & \multicolumn{1}{c|}{0.028} & 0.029 & \multicolumn{1}{c|}{2.252} & 1.077 & \multicolumn{1}{c|}{0.030} & 0.051 & 0.255 & \textbf{2.684} \\
S & \cite{poggi2020uncertainty}-Boot+Log & 8 & 8 & 0.085 & {\ul 3.777} & 0.913 & \multicolumn{1}{c|}{{\ul 0.020}} & \textbf{0.038} & \multicolumn{1}{c|}{0.807} & 2.455 & \multicolumn{1}{c|}{\textbf{0.018}} & \textbf{0.063} & - & - \\
S & \cite{poggi2020uncertainty}-Boot+Self & 9 & 8 & 0.085 & 3.793 & 0.914 & \multicolumn{1}{c|}{0.023} & 0.035 & \multicolumn{1}{c|}{1.646} & 1.628 & \multicolumn{1}{c|}{0.021} & 0.058 & 0.398 & 4.288 \\
S & \cite{poggi2020uncertainty}-Snap+Log & 1 & 8 & \textbf{0.083} & 3.833 & 0.914 & \multicolumn{1}{c|}{0.021} & {\ul 0.037} & \multicolumn{1}{c|}{0.891} & 2.426 & \multicolumn{1}{c|}{\textbf{0.018}} & {\ul 0.061} & - & - \\
S & \cite{poggi2020uncertainty}-Snap+Self & 2 & 8 & 0.086 & 3.859 & 0.912 & \multicolumn{1}{c|}{0.023} & 0.035 & \multicolumn{1}{c|}{1.710} & 1.623 & \multicolumn{1}{c|}{0.023} & 0.058 & 0.282 & 3.025 \\
S & Ours & 1 & 1 & {\ul 0.084} & 3.834 & \textbf{0.916} & \multicolumn{1}{c|}{0.023} & 0.033 & \multicolumn{1}{c|}{{\ul 0.661}} & {\ul 2.655} & \multicolumn{1}{c|}{0.023} & 0.055 & {\ul 0.075} & 3.540 \\
S & Ours-self & 2 & 1 & {\ul 0.084} & 3.792 & 0.914 & \multicolumn{1}{c|}{\textbf{0.018}} & \textbf{0.038} & \multicolumn{1}{c|}{\textbf{0.349}} & \textbf{2.924} & \multicolumn{1}{c|}{{\ul 0.019}} & 0.060 & \textbf{0.072} & 3.068 \\ \hline
MS & \cite{godard2019digging} & 1 & 1 & 0.084 & 3.739 & {\ul 0.918} & \multicolumn{1}{c|}{-} & - & \multicolumn{1}{c|}{-} & - & \multicolumn{1}{c|}{-} & - & - & - \\
MS & \cite{godard2019digging}-Post & 1 & 2 & \textbf{0.082} & \textbf{3.666} & \textbf{0.919} & \multicolumn{1}{c|}{0.036} & 0.018 & \multicolumn{1}{c|}{2.498} & 0.655 & \multicolumn{1}{c|}{0.044} & 0.031 & 0.290 & {\ul 2.974} \\
MS & \cite{poggi2020uncertainty}-Repr & 1 & 1 & 0.084 & 3.828 & 0.913 & \multicolumn{1}{c|}{0.046} & 0.010 & \multicolumn{1}{c|}{2.662} & 0.635 & \multicolumn{1}{c|}{0.062} & 0.018 & - & - \\
MS & \cite{poggi2020uncertainty}-Log & 1 & 1 & {\ul 0.083} & 3.790 & 0.916 & \multicolumn{1}{c|}{0.028} & 0.029 & \multicolumn{1}{c|}{1.714} & 1.562 & \multicolumn{1}{c|}{0.028} & 0.050 & - & - \\
MS & \cite{poggi2020uncertainty}-Self & 2 & 1 & {\ul 0.083} & 3.682 & \textbf{0.919} & \multicolumn{1}{c|}{{\ul 0.022}} & 0.033 & \multicolumn{1}{c|}{1.654} & 1.515 & \multicolumn{1}{c|}{{\ul 0.023}} & 0.052 & 0.083 & 3.686 \\
MS & \cite{poggi2020uncertainty}-Drop & 1 & 8 & 0.172 & 5.885 & 0.679 & \multicolumn{1}{c|}{0.103} & -0.027 & \multicolumn{1}{c|}{7.114} & -2.580 & \multicolumn{1}{c|}{0.303} & -0.081 & 5.547 & 62.54 \\
MS & \cite{poggi2020uncertainty}-Boot & 8 & 8 & 0.086 & 3.787 & 0.910 & \multicolumn{1}{c|}{0.028} & 0.030 & \multicolumn{1}{c|}{2.269} & 0.985 & \multicolumn{1}{c|}{0.034} & 0.049 & 0.564 & 5.898 \\
MS & \cite{poggi2020uncertainty}-Snap & 1 & 8 & 0.085 & 3.806 & 0.914 & \multicolumn{1}{c|}{0.029} & 0.028 & \multicolumn{1}{c|}{2.245} & 1.029 & \multicolumn{1}{c|}{0.033} & 0.047 & 0.254 & \textbf{2.706} \\
MS & \cite{poggi2020uncertainty}-Boot+Log & 8 & 8 & 0.086 & 3.771 & 0.911 & \multicolumn{1}{c|}{0.030} & 0.028 & \multicolumn{1}{c|}{1.962} & 1.282 & \multicolumn{1}{c|}{0.032} & 0.051 & - & - \\
MS & \cite{poggi2020uncertainty}-Boot+Self & 9 & 8 & 0.085 & 3.704 & 0.915 & \multicolumn{1}{c|}{0.023} & 0.033 & \multicolumn{1}{c|}{1.688} & 1.494 & \multicolumn{1}{c|}{{\ul 0.023}} & {\ul 0.056} & 0.355 & 3.842 \\
MS & \cite{poggi2020uncertainty}-Snap+Log & 1 & 8 & 0.084 & 3.828 & 0.914 & \multicolumn{1}{c|}{0.030} & 0.027 & \multicolumn{1}{c|}{2.032} & 1.272 & \multicolumn{1}{c|}{0.032} & 0.048 & - & - \\
MS & \cite{poggi2020uncertainty}-Snap+Self & 2 & 8 & 0.085 & 3.715 & 0.916 & \multicolumn{1}{c|}{0.023} & {\ul 0.034} & \multicolumn{1}{c|}{1.684} & 1.510 & \multicolumn{1}{c|}{{\ul 0.023}} & 0.055 & 0.276 & 2.979 \\
MS & Ours & 1 & 1 & 0.084 & 3.806 & 0.915 & \multicolumn{1}{c|}{0.027} & 0.029 & \multicolumn{1}{c|}{{\ul 0.840}} & {\ul 2.436} & \multicolumn{1}{c|}{0.029} & 0.049 & \textbf{0.077} & 3.573 \\
MS & Ours-self & 2 & 1 & \textbf{0.082} & {\ul 3.667} & \textbf{0.919} & \multicolumn{1}{c|}{\textbf{0.016}} & \textbf{0.039} & \multicolumn{1}{c|}{\textbf{0.293}} & \textbf{2.859} & \multicolumn{1}{c|}{\textbf{0.014}} & \textbf{0.061} & {\ul 0.078} & 3.528 \\ \hline
\end{tabular}
\caption{Results of monocular video supervision only (M), stereo supervision only (S), and both monocular video and stereo supervision (MS) trainings of our MonoProb methods with and without self-distillation compared to methods that require one or more inferences to provide an uncertainty estimate.}
\label{tab:all_comparisons}
\end{table*}

\section{Complementary qualitative results}

\begin{figure*}[t]
    \centering
    \includegraphics[width=1.01\textwidth]{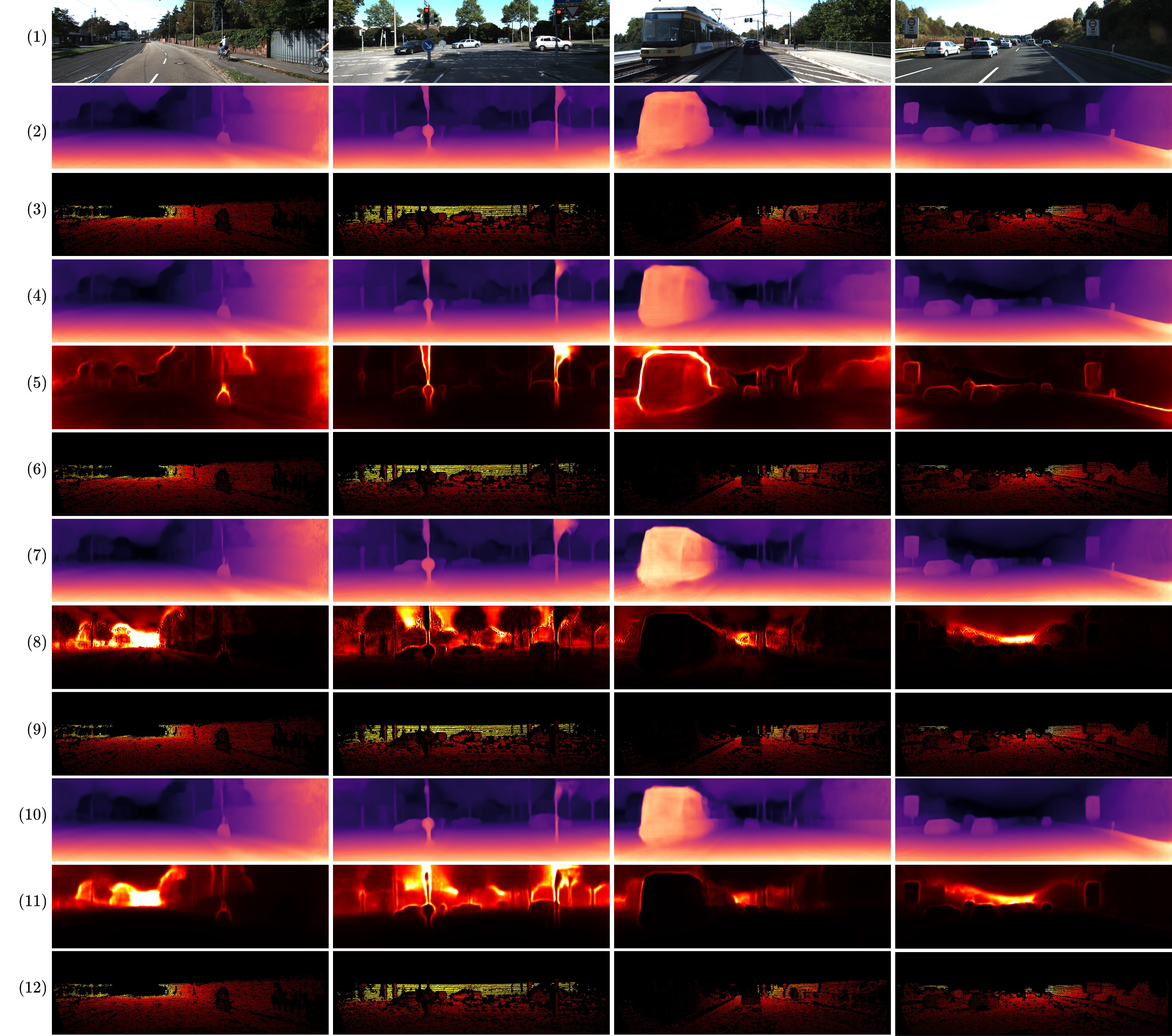}
    \caption{Qualitative results of monocular trainings on KITTI \cite{kitti}. From top to bottom, (1) input image, (2) depth map from \cite{godard2019digging}, (3) error map from \cite{godard2019digging}, (4) depth map from \cite{poggi2020uncertainty}-Self, (5) uncertainty map from \cite{poggi2020uncertainty}-Self, (6) error map from \cite{poggi2020uncertainty}-Self, (7) depth map from our MonoProb without self-distillation, (8) uncertainty map from our MonoProb without self-distillation, (9) error map from our MonoProb without self-distillation, (10) depth map from our self-distilled MonoProb, (11) uncertainty map from our self-distilled MonoProb, (12) error map from our self-distilled MonoProb.}
    \label{fig:visu_kitti_depth}
\end{figure*}

\begin{figure*}[t]
    \centering
    \includegraphics[width=1.01\textwidth]{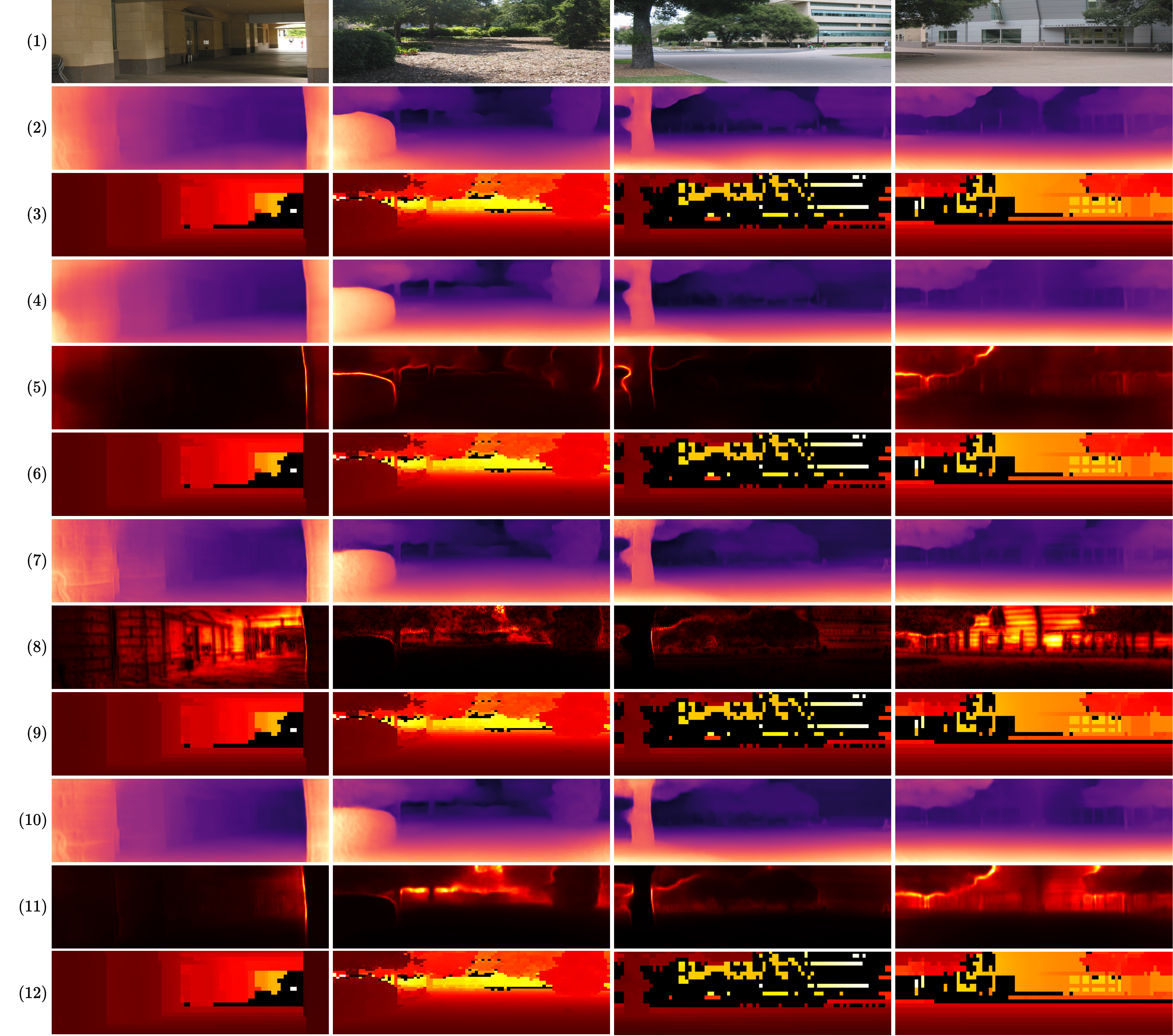}
    \caption{Qualitative results of monocular trainings on Make3D \cite{make3d}. From top to bottom, (1) input image, (2) depth map from \cite{godard2019digging}, (3) error map from \cite{godard2019digging}, (4) depth map from \cite{poggi2020uncertainty}-Self, (5) uncertainty map from \cite{poggi2020uncertainty}-Self, (6) error map from \cite{poggi2020uncertainty}-Self, (7) depth map from our MonoProb without self-distillation, (8) uncertainty map from our MonoProb without self-distillation, (9) error map from our MonoProb without self-distillation, (10) depth map from our self-distilled MonoProb, (11) uncertainty map from our self-distilled MonoProb, (12) error map from our self-distilled MonoProb.}
    \label{fig:visu_make3d}
\end{figure*}

\begin{figure*}[h]
    \centering
    \includegraphics[width=1.01\textwidth]{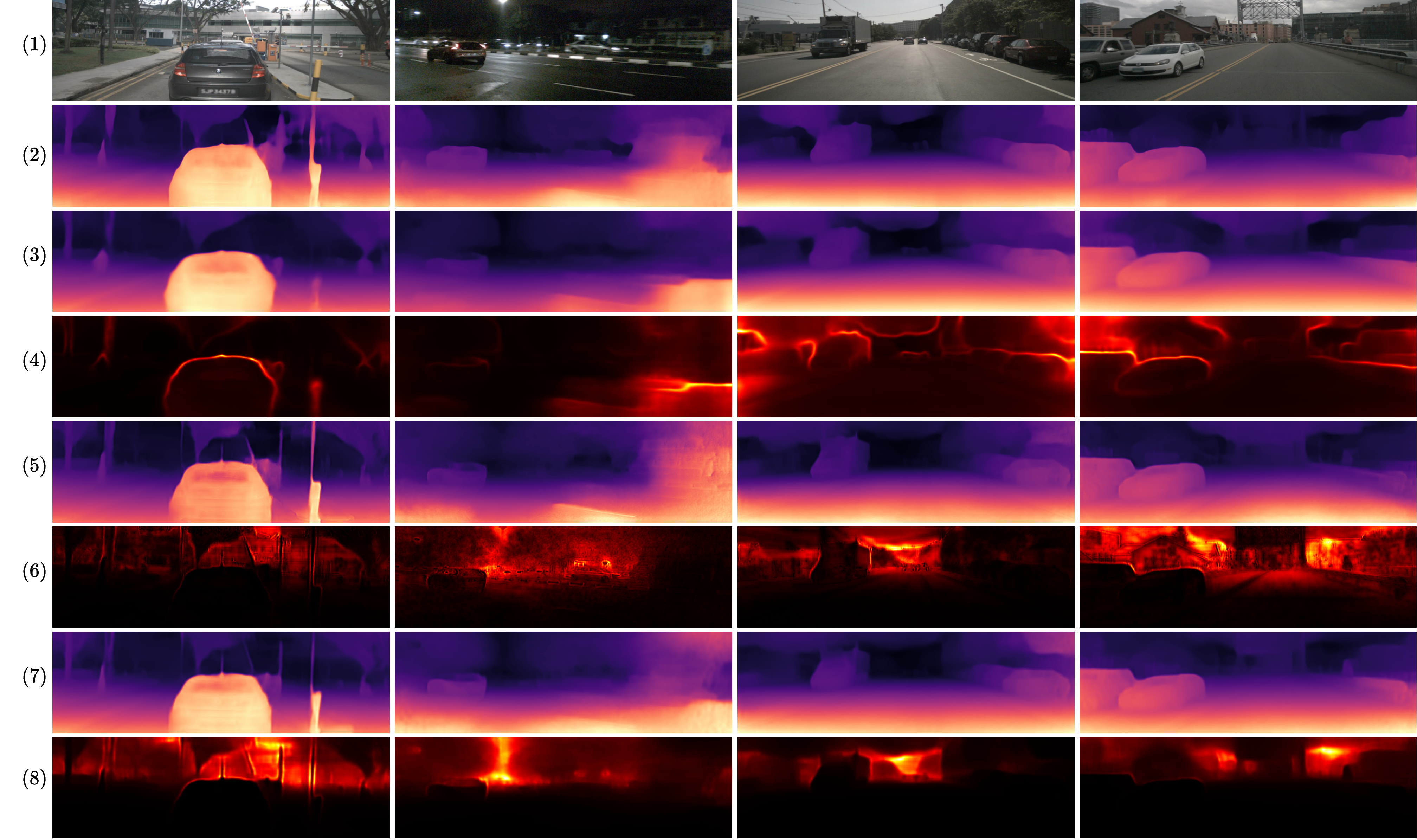}
    \caption{Qualitative results of monocular trainings on Nuscenes \cite{nuscenes}. From top to bottom, (1) input image, (2) depth map from \cite{godard2019digging}, (3) depth map from \cite{poggi2020uncertainty}-Self, (4) uncertainty map from \cite{poggi2020uncertainty}-Self, (5) depth map from our MonoProb without self-distillation, (6) uncertainty map from our  MonoProb without self-distillation, (7) depth map from our self-distilled MonoProb, (8) uncertainty map from our self-distilled MonoProb. We do not provide error maps because the high sparsity of the ground truth makes them unreadable.}
    \label{fig:visu_nuscenes}
\end{figure*}

We provide qualitative results on KITTI \cite{kitti} in \cref{fig:visu_kitti_depth} Make3D \cite{make3d} in \cref{fig:visu_make3d} and Nuscenes \cite{nuscenes} in \cref{fig:visu_nuscenes}.

\newpage
\begin{equation*}
\end{equation*}
\begin{equation*}
\end{equation*}
\begin{equation*}
\end{equation*}
\begin{equation*}
\end{equation*}
\begin{equation*}
\end{equation*}
\begin{equation*}
\end{equation*}
\begin{equation*}
\end{equation*}
\begin{equation*}
\end{equation*}
\begin{equation*}
\end{equation*}
\begin{equation*}
\end{equation*}
\begin{equation*}
\end{equation*}
\begin{equation*}
\end{equation*}
\begin{equation*}
\end{equation*}
\begin{equation*}
\end{equation*}
\begin{equation*}
\end{equation*}
\begin{equation*}
\end{equation*}
\begin{equation*}
\end{equation*}
\begin{equation*}
\end{equation*}
\begin{equation*}
\end{equation*}
\begin{equation*}
\end{equation*}
\begin{equation*}
\end{equation*}
\begin{equation*}
\end{equation*}
\begin{equation*}
\end{equation*}
\begin{equation*}
\end{equation*}
\begin{equation*}
\end{equation*}
\begin{equation*}
\end{equation*}
\begin{equation*}
\end{equation*}
\begin{equation*}
\end{equation*}
\begin{equation*}
\end{equation*}
\begin{equation*}
\end{equation*}
\begin{equation*}
\end{equation*}
\begin{equation*}
\end{equation*}
\begin{equation*}
\end{equation*}
\begin{equation*}
\end{equation*}
\begin{equation*}
\end{equation*}
\begin{equation*}
\end{equation*}
\begin{equation*}
\end{equation*}
\begin{equation*}
\end{equation*}
\begin{equation*}
\end{equation*}
\begin{equation*}
\end{equation*}
\begin{equation*}
\end{equation*}
\begin{equation*}
\end{equation*}
\begin{equation*}
\end{equation*}
\begin{equation*}
\end{equation*}
\begin{equation*}
\end{equation*}
\begin{equation*}
\end{equation*}
\begin{equation*}
\end{equation*}
\begin{equation*}
\end{equation*}
\begin{equation*}
\end{equation*}
\begin{equation*}
\end{equation*}
\begin{equation*}
\end{equation*}
\begin{equation*}
\end{equation*}
\begin{equation*}
\end{equation*}
\begin{equation*}
\end{equation*}
\begin{equation*}
\end{equation*}
\begin{equation*}
\end{equation*}
\begin{equation*}
\end{equation*}
\begin{equation*}
\end{equation*}
\begin{equation*}
\end{equation*}
\begin{equation*}
\end{equation*}
\begin{equation*}
\end{equation*}
\begin{equation*}
\end{equation*}
\begin{equation*}
\end{equation*}
\begin{equation*}
\end{equation*}
\begin{equation*}
\end{equation*}
\begin{equation*}
\end{equation*}
\begin{equation*}
\end{equation*}
\begin{equation*}
\end{equation*}
\begin{equation*}
\end{equation*}
\begin{equation*}
\end{equation*}
\begin{equation*}
\end{equation*}
\begin{equation*}
\end{equation*}
\begin{equation*}
\end{equation*}
\begin{equation*}
\end{equation*}
\begin{equation*}
\end{equation*}
\begin{equation*}
\end{equation*}
\begin{equation*}
\end{equation*}
\begin{equation*}
\end{equation*}
\begin{equation*}
\end{equation*}
\begin{equation*}
\end{equation*}
\begin{equation*}
\end{equation*}
\begin{equation*}
\end{equation*}
\begin{equation*}
\end{equation*}
\begin{equation*}
\end{equation*}
\begin{equation*}
\end{equation*}
\begin{equation*}
\end{equation*}
\begin{equation*}
\end{equation*}
\begin{equation*}
\end{equation*}
\begin{equation*}
\end{equation*}
\begin{equation*}
\end{equation*}

%%%%%%%%% REFERENCES
{\small
\bibliographystyle{ieee_fullname}
\bibliography{main}
}

\end{document}